\pdfoutput=1
\documentclass{article} 
\usepackage[preprint]{colm2026_conference}

\usepackage{microtype}
\usepackage{hyperref}
\usepackage{url}
\usepackage{booktabs}
\usepackage{multirow}

\usepackage{lineno}
\definecolor{darkblue}{rgb}{0, 0, 0.5}
\hypersetup{colorlinks=true, citecolor=darkblue, linkcolor=darkblue, urlcolor=darkblue}
\usepackage{makecell}
\usepackage{graphicx}

\usepackage{amsmath}
\usepackage{amssymb}
\usepackage{amsfonts}
\usepackage{enumitem}
\usepackage[most]{tcolorbox}
\usepackage{colortbl}
\usepackage{wrapfig}
\definecolor{firstbg}{RGB}{214,234,248}
\definecolor{secondbg}{RGB}{235,235,235}

\title{Faithful GRPO: Improving Visual Spatial Reasoning in Multimodal Language Models via Constrained Policy Optimization}

\author{
\begin{tabular}{cc}
\begin{tabular}[t]{c}
Sai Srinivas Kancheti\thanks{Equal contribution}\thanks{Work done at Microsoft Research} \\
CSE Dept., IIT Hyderabad \\
Hyderabad, India \\
\texttt{cs21resch01004@iith.ac.in}
\end{tabular} &
\begin{tabular}[t]{c}
Aditya Kanade\footnotemark[1] \\
Microsoft Research \\
Bengaluru, India \\
\texttt{kanade850@gmail.com}
\end{tabular} \\
\\[1em]
\begin{tabular}[t]{c}
Rohit Sinha \\
CSE Dept., IIT Hyderabad \\
Hyderabad, India \\
\texttt{rohit.sinha@prjt.cse.iith.ac.in}
\end{tabular} &
\begin{tabular}[t]{c}
Vineeth N Balasubramanian\thanks{Corresponding author} \\
Microsoft Research \\
Bengaluru, India \\
\texttt{vineeth.nb@microsoft.com}
\end{tabular} \\
\\[1em]
\multicolumn{2}{c}{
\begin{tabular}[t]{c}
Tanuja Ganu\footnotemark[3] \\
Microsoft Research \\
Bengaluru, India \\
\texttt{tanuja.ganu@microsoft.com}
\end{tabular}
} \\
\end{tabular}
}

\begin{document}

\ifcolmsubmission
\linenumbers
\fi

\maketitle

\begin{abstract}
Multimodal reasoning models (MRMs) trained with reinforcement learning with verifiable rewards (RLVR) show improved accuracy on visual reasoning benchmarks. However, we observe that accuracy gains often come at the cost of reasoning quality: generated Chain-of-Thought (CoT) traces are frequently inconsistent with the final answer and poorly grounded in the visual evidence. We systematically study this phenomenon across seven challenging real-world spatial reasoning benchmarks and find that it affects contemporary MRMs such as ViGoRL-Spatial, TreeVGR as well as our own models trained with standard Group Relative Policy Optimization (GRPO). We characterize CoT reasoning quality along two complementary axes: \emph{logical consistency} (does the CoT entail the final answer?) and \emph{visual grounding} (does each reasoning step accurately describe objects, attributes, and spatial relationships in the image?).
To address this, we propose Faithful GRPO (FGRPO), a variant of GRPO that enforces consistency and grounding as constraints via Lagrangian dual ascent.
FGRPO incorporates batch-level consistency and grounding constraints into the advantage computation within a group, adaptively adjusting the relative importance of constraints during optimization.
We evaluate FGRPO on Qwen2.5-VL-7B and 3B backbones across seven spatial datasets. Our results show that FGRPO substantially improves reasoning quality, reducing the inconsistency rate from 24.5\% to 1.7\% and improving visual grounding scores by +13\%. It also improves final answer accuracy over simple GRPO, demonstrating that faithful reasoning enables better answers.
\end{abstract}


\section{Introduction}
\label{sec:intro}

Multimodal Large Language Models~\citep{qwen2.5vl_techreport,InernVL3,llavaonevision} (MLMs) have achieved strong performance on visual understanding tasks~\citep{Yue2024MMMUProAM,Liu2023MMBenchIY,Lu2023MathVistaEM}, substantially enhanced by Chain-of-Thought (CoT) reasoning~\citep{cot1,cot2}. Reinforcement Learning with Verifiable Rewards (RLVR)~\citep{TULU3,trung2024reft}, as demonstrated by DeepSeek-R1~\citep{DeepSeekAI2025DeepSeekR1IR}, has emerged as the dominant paradigm for training reasoning capabilities via a two-stage recipe of supervised finetuning (SFT) followed by Group Relative Policy Optimization (GRPO)~\citep{grpo}. Following this paradigm, we train a task-reward model (GRPO-T) on diverse spatial reasoning data (\S\ref{subsec:preliminaries}), which shows accuracy gains over the Qwen2.5-VL backbone ($65.17\%$ vs Qwen2.5-VL-7B backbone $64.17\%$ from Table~\ref{tab:main_results} ).

\begin{figure*}[t]
    \centering
    \includegraphics[width=0.8\textwidth]{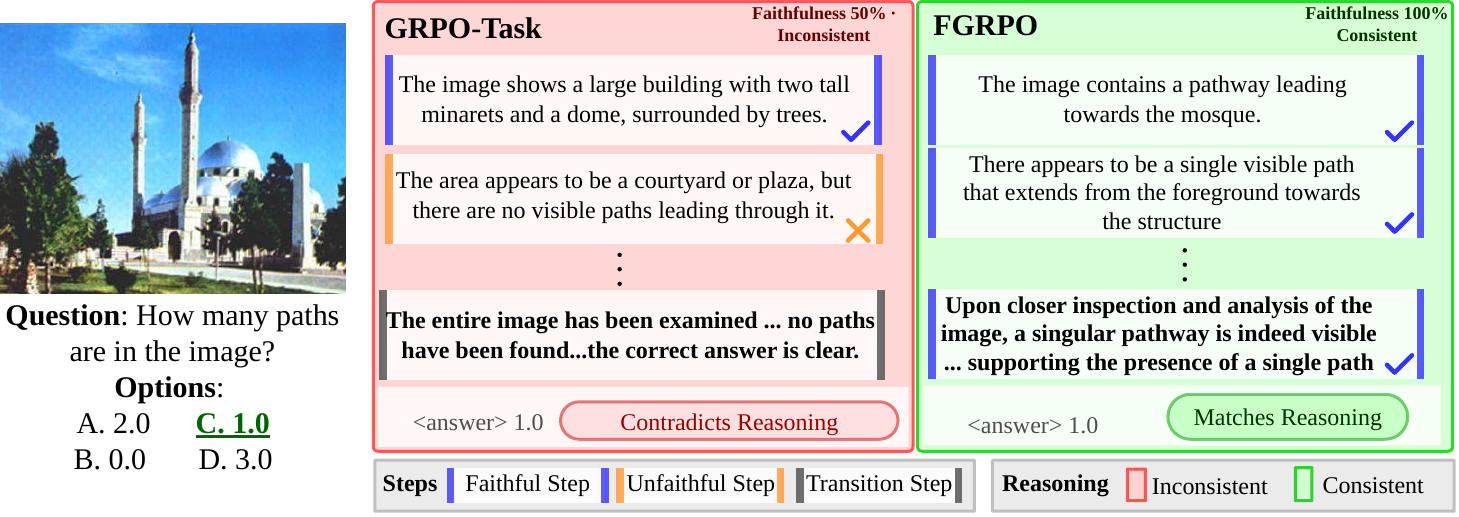}\\[0.5em]
    \includegraphics[width=0.8\textwidth]{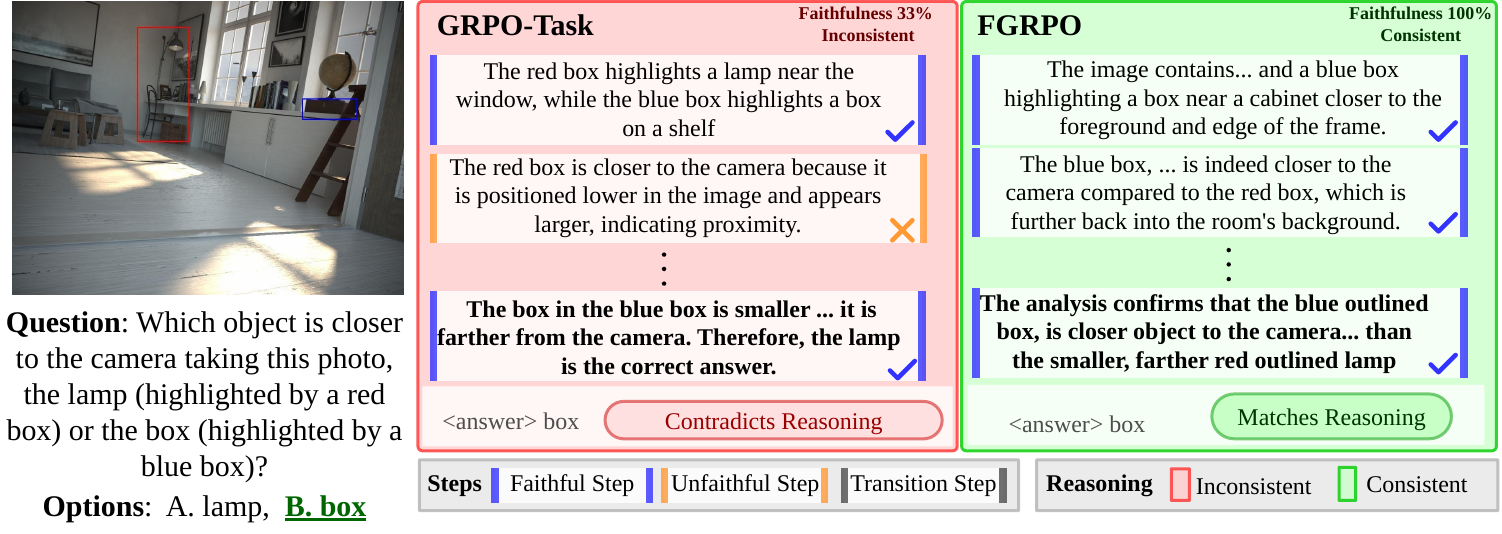}
    \caption{\textbf{Unfaithful reasoning masked by correct answers.} Both models answer correctly, but only FGRPO reasons faithfully. \textbf{Left:} The GRPO-Task model incorrectly claims there are no visible paths, contradicting its own answer of ``1.0'' (50\% faithfulness, inconsistent). \textbf{Right:} The GRPO-Task model reasons toward ``lamp'' but answers ``box'' (33\% faithfulness, inconsistent). FGRPO produces visually grounded reasoning in both cases (100\% faithfulness, consistent). Sentences are color-coded: blue are grounded and orange are ungrounded.}
    \label{fig:contrastive_examples}
\end{figure*}

However, accuracy alone is an incomplete measure of reasoning ability. As shown in Figure~\ref{fig:contrastive_examples}, we observe pervasive reasoning failures in the traces generated by GRPO-T. A model that generates the correct answer despite flawed, contradictory, or visually ungrounded reasoning may have learned to exploit shortcuts~\citep{shortcut} and biases to answer the question, making it untrustworthy for real-world use. We identify two distinct failure modes in CoT traces generated by trained MRMs: \textbf{(i) Logical Inconsistency}, where the reasoning trace argues toward one conclusion, but the model then abruptly flips its prediction to a different final answer (Figure~\ref{fig:contrastive_examples}, right: the GRPO-T model reasons toward `lamp' but answers `box'); and \textbf{(ii) Visual Ungroundedness}, where individual reasoning steps describe objects, attributes or spatial relationships that are inaccurate with respect to the visual content (Figure~\ref{fig:contrastive_examples}, left: the GRPO-T model claims there are no visible paths). A trace can be logically consistent yet visually ungrounded, or well-grounded yet inconsistent with its own answer. We provide additional contrastive examples spanning diverse spatial tasks in Appendix~\ref{appx:qualitative}.

We address these failure modes by defining verifiable reward signals for both consistency and grounding (\S\ref{subsec:diagnosing}). Consistency is treated as a binary reward via an LLM judge. For visual grounding, we combine a per-sentence semantic grounding reward scored by a VLM judge with a spatial grounding reward based on IoU matching of predicted bounding boxes to ground-truth regions. Each reward directly targets a specific failure mode: the consistency reward addresses logical inconsistency, while the grounding rewards address visual ungroundedness.

We then propose \textbf{Faithful GRPO (FGRPO)}, a constrained variant of GRPO that maximizes task accuracy subject to minimum thresholds on consistency and grounding rewards, enforced via Lagrangian dual ascent~\citep{altman1999constrained,tessler2019rcpo}. In our formulation, the Lagrange multipliers adaptively increase pressure on violated constraints and decrease it on satisfied ones, removing the need for manual reward weight tuning. FGRPO treats consistency and grounding as \emph{prerequisites} for trustworthy visual reasoning, making them natural candidates for constraints that need to be satisfied. 


We evaluate FGRPO on Qwen2.5-VL-7B and 3B backbones across seven spatial reasoning datasets. Our results demonstrate that FGRPO not only substantially improves reasoning quality by reducing inconsistency rates and improving visual grounding scores, but also improves final answer accuracy over standard GRPO. 

\noindent We summarize our contributions below:
\begin{itemize}[nosep,leftmargin=*]
    \item We characterize reasoning quality degradation in RLVR-trained MRMs along two axes: logical consistency and visual grounding. We observe this degradation in contemporary MRMs and in our own models trained with standard GRPO.
    \item We define verifiable reward signals for trustworthy visual reasoning, and propose Faithful GRPO (FGRPO), which treats consistency and grounding rewards as constraints to be satisfied for visual reasoning.
    \item We evaluate FGRPO on two backbones across seven spatial datasets and show that it improves both accuracy and reasoning quality, demonstrating that faithful reasoning and accurate answers are complementary objectives.
\end{itemize}
\section{Related Work}
\label{sec:related_work}

\noindent\textbf{RLVR for Multimodal Reasoning.}
Reinforcement Learning with Verifiable Rewards (RLVR)~\citep{TULU3,trung2024reft} optimizes models on tasks with automatically checkable outcomes, sidestepping the need for learned reward models. DeepSeek-R1~\citep{DeepSeekAI2025DeepSeekR1IR} demonstrates that a two-stage pipeline -- SFT on CoT traces followed by GRPO~\citep{grpo} with verifiable rewards yields strong reasoning performance across domains~\citep{Wen2025ReinforcementLW}. A growing body of work extends this paradigm to multimodal reasoning. Vision-R1~\citep{visionr1} constructs large-scale CoT SFT data and introduces progressive thinking suppression during RL. VL-Rethinker~\citep{vlrethinker} encourages self-reflection via forced rethinking triggers and selective sample replay. R1-OneVision~\citep{r1onevision} converts visual inputs into structured textual representations before applying language-only reasoning. TreeVGR~\citep{treevgr} and ViGoRL~\citep{vigorl} focus on visually grounded reasoning: TreeVGR supervises both localization and reasoning with dual IoU-based rewards, while ViGoRL grounds every reasoning step with image coordinates via MCTS-generated point-grounded CoT traces. Our work builds on the same two-stage paradigm but departs from prior MRMs in two ways: (i)~we define verifiable reward signals for consistency and visual grounding that have not been used as trainable rewards in prior work, and (ii)~we treat these signals as hard constraints rather than reward terms, addressing the accuracy--faithfulness tradeoff that affects existing MRMs.

\noindent\textbf{Constrained Policy Optimization.}
Constrained Markov Decision Processes (CMDPs)~\citep{altman1999constrained} provide a principled framework for optimizing a primary objective subject to auxiliary constraints. Lagrangian relaxation is a standard tool for solving CMDPs, employed in methods such as CPO~\citep{achiam2017cpo}, RCPO~\citep{tessler2019rcpo}, PID Lagrangian methods~\citep{stooke2020responsive}, and first-order constrained optimization in policy space~\citep{zhang2020focops}. These approaches have been applied to safe exploration~\citep{ray2019benchmarking}, LLM alignment~\citep{dai2024safe_rlhf}, and recently to safety constraints for vision-language-action models~\citep{zhang2025safevla}. Concurrent to our work, MO-GRPO~\citep{mogrpo} identifies that standard GRPO suffers from reward hacking under multiple objectives because within-group normalization allows high-variance rewards to dominate, and proposes scalarized per-objective advantages to equalize their influence. GDPO~\citep{gdpo} similarly advocates decoupled normalization and introduces conditional objectives that activate reward terms only once prerequisite scores exceed a minimum threshold. FGRPO shares the decoupled normalization strategy with these methods but differs in its use of Lagrangian dual ascent to enforce constraint thresholds rather than fixed or conditional weights, and in its application to enforcing reasoning quality (consistency and visual grounding) in multimodal RL.



\section{Methodology}
\label{sec:methodology}

We begin by describing the standard GRPO training setup for multimodal reasoning in \S~\ref{subsec:preliminaries}.
We then define two complementary axes along which reasoning quality can be measured (\S\ref{subsec:diagnosing}) and describe our reward formulation in detail. In \S~\ref{subsec:fgrpo}, we present Faithful GRPO, which treats consistency and visual grounding as constraints enforced via Lagrangian dual ascent. An overview of the FGRPO training pipeline is shown in Figure~\ref{fig:training_flow}.

\subsection{Preliminaries}
\label{subsec:preliminaries}

\paragraph{Backbone and Two-Stage Training.}
We build on the Qwen2.5-VL-7B-Instruct backbone~\citep{qwen2.5vl_techreport} and follow the two-stage training paradigm of DeepSeek-R1~\citep{DeepSeekAI2025DeepSeekR1IR}. In stage one, we perform SFT on curated CoT data to imbue the backbone with baseline spatial reasoning capability. We create CoT data from a strong visual teacher (Qwen2.5-VL-72B-Instruct) using Monte Carlo Tree Search (MCTS)~\citep{mcts1,mcts2}, which generates diverse, high-quality trajectories including synthetic backtracking for self-correction (Appendix~\ref{appx:mcts_procedure}). We curate approximately 45K CoT traces from three seed datasets: SAT~\citep{ray2025sat}, VGR~\citep{wang2025vgrvisualgroundedreasoning}, and VisCoT~\citep{shao2024visual} that collectively span diverse real-world spatial questions and image sources including COCO, GQA, OpenImages, and Flickr30k. In stage two, the SFT checkpoint is finetuned with reinforcement learning using GRPO~\citep{grpo} on a curated RL dataset of 49K samples. We employ difficulty-based filtering~\citep{less,Yu2025DAPOAO} to select samples of intermediate difficulty and include 13K samples from TreeVGR-RL-37K~\citep{treevgr} for bounding-box supervision diversity. We present the full training hyperparameters, data curation details, and the MCTS procedure in Appendix~\ref{appx:training_details}.

\paragraph{GRPO.}
Let $\pi_\theta$ denote the policy, which autoregressively generates a response $o = (s_1, s_2, \ldots, s_T, a)$ conditioned on a multimodal prompt $x$, where $s_t$ denotes a reasoning step and $a$ the final answer. Given a batch of $N$ prompts, GRPO~\citep{grpo} generates $G$ rollouts $\{o^j_i\}_{i=1}^G$ for each prompt $x_j$ and computes per-rollout scalar rewards $r^j_i = R(o^j_i)$. The advantage is obtained by normalizing within each group:
$\hat{A}^j_i = ({r^j_i - \mu_j})/({\sigma_j + \epsilon}),$
where $\mu_j$ and $\sigma_j$ are the mean and standard deviation of rewards within group $j$. The policy is updated via the clipped surrogate objective with a KL penalty~\citep{grpo}.

\paragraph{Task Reward.}
We use the standard task reward that combines format adherence and answer accuracy: $R_{\mathrm{task}}(o) = 0.5 \cdot R_{\mathrm{acc}}(o) + 0.5 \cdot R_{\mathrm{fmt}}(o),$ 
where $R_{\mathrm{fmt}}(o) = 1$ if $o$ follows the \texttt{<think>...<\!/think><answer>...<\!/answer>} format and $0$ otherwise, and $R_{\mathrm{acc}}(o) = 1$ if the answer matches the ground truth and $0$ otherwise. Since our RL dataset is formulated as MCQs, we perform exact matching after stripping punctuation. 

\subsection{Consistency and Grounding Rewards}
\label{subsec:diagnosing}

Training with $R_{\mathrm{task}}$ improves answer accuracy, but as discussed in \S\ref{sec:intro}, this improvement frequently comes at the expense of reasoning quality. We now define verifiable reward signals that capture the two failure modes identified above -- logical inconsistency and visual ungroundedness, enabling them to be used as training objectives.

\noindent\textbf{Consistency Reward.}
We define the \emph{consistency reward} $R_C(o) \in \{0, 1\}$ as a binary signal indicating whether the CoT reasoning trace logically entails the final answer. Given a response $o$ with reasoning trace $\mathcal{T}$ (the text within \texttt{<think>} tags) and final answer $\mathcal{A}$ (the text within \texttt{<answer>} tags), we prompt a text-only LLM judge to determine whether $\mathcal{A}$ follows logically from $\mathcal{T}$: $R_C(o) = \mathrm{LLM\text{-}Judge}(\mathcal{T}, \mathcal{A}) \in \{0, 1\},$
where the judge outputs $1$ (consistent) if the reasoning's conclusion matches the final answer, and $0$ (inconsistent) otherwise. The judge evaluates only textual logical coherence, ignoring visual correctness. A trace that reasons incorrectly about the image but answers in accordance with its own reasoning is scored as consistent. We mask this reward to samples where $R_{\mathrm{acc}}(o) > 0$, since consistency is only meaningful when the model has produced a non-trivial answer. The full judge prompt is provided in Appendix~\ref{appx:prompts}. 

\noindent\textbf{Visual Grounding Rewards.}
We define two complementary reward signals that capture whether the reasoning trace is anchored in the visual evidence.

\noindent\emph{Semantic grounding reward} $R_S(o) \in [0, 1]$ measures whether individual reasoning steps accurately describe the objects, attributes, and spatial relationships visible in the image. We decompose the reasoning trace into sentences $\{s_1, \ldots, s_K\}$, filter out trivial non-visual sentences (meta-reasoning, planning, hedging), and score each remaining sentence via a VLM judge that receives the image, the question, and the reasoning context: $R_S(o) = \frac{1}{K'} \sum_{k=1}^{K'} \hat{s}_k, \quad \hat{s}_k = \mathrm{VLM\text{-}Judge}(I, Q, s_{\leq k}) \in \{0, 1\},$
where $K'$ is the number of non-trivial scored sentences and $\hat{s}_k = 1$ if the sentence is classified as CORRECT (the visual claim is accurate) and $0$ if INCORRECT. Sentences classified as SKIP (no visual claim) are excluded. The VLM judge checks entity grounding, attribute verification, spatial relationship accuracy, and bounding-box content validity. We mask this reward to samples where $R_{\mathrm{acc}}(o) > 0$. Details of the sentence decomposition, trivial sentence filtering, and the VLM judge prompt are provided in Appendix~\ref{appx:semantic_grounding_details}.

\noindent\emph{Spatial grounding reward} $R_G(o) \in [0, 1]$ measures whether bounding-box coordinates generated in the reasoning trace correspond to the correct image regions. For models that produce bounding boxes $\hat{B}(o) = \{\hat{b}_i\}_{i=1}^N$ within \texttt{<bbox>} tags, we compute $R_G(o)$ via Hungarian matching~\citep{hungarian} against ground-truth boxes $B = \{b_j\}_{j=1}^M$ using Complete IoU (CIoU)~\citep{ciou}:
\begin{equation}
\label{eq:spatial_grounding}
R_G(o) = \frac{1}{\max(N, M)} \sum_{i=1}^{N} \mathrm{CIoU}(\hat{b}_i, b_{\sigma(i)}),
\end{equation}
where $\sigma$ denotes the optimal assignment. CIoU incorporates intersection-over-union, center distance, and aspect ratio penalties. This reward is masked to training samples with bounding-box annotations (VGR, TreeVGR datasets).

\paragraph{Empirical Motivation.}
We evaluate our task-reward trained model GRPO-T on seven spatial benchmarks using these reward signals as diagnostic metrics. Despite achieving 65.2\% average accuracy, GRPO-T exhibits an inconsistency rate of 26.1\%---over one in four samples has reasoning that contradicts the final prediction. Its mean semantic grounding score is 72.7\%, indicating that over a quarter of visual claims in the reasoning traces are inaccurate with respect to the image. We observe similar degradation across five contemporary MRM baselines (Table~\ref{tab:main_results}, Figure~\ref{fig:faithfulness_consistency}). Naively adding $R_C$ and $R_G$ as reward terms does not resolve this: additive shaping reduces inconsistency but sacrifices accuracy, while multiplicative gating degrades both (Table~\ref{tab:reward_shaping}). These findings motivate treating consistency and grounding not as soft reward terms but as hard constraints.
\begin{figure*}[t]
    \centering
    \includegraphics[width=\textwidth]{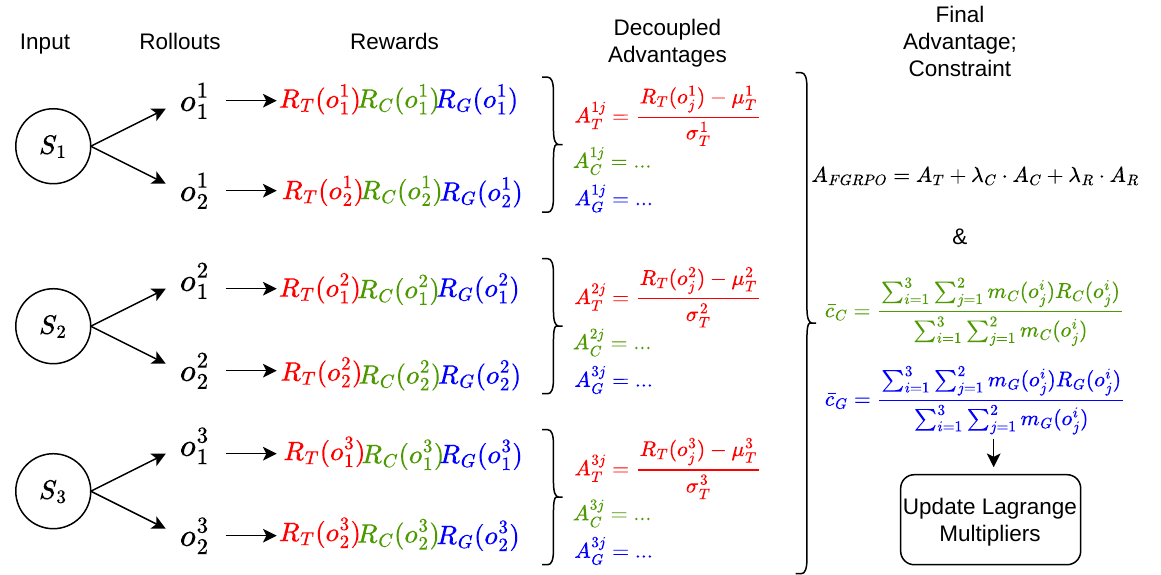}
    \caption{\textbf{Overview of the FGRPO training pipeline.} We show advantage computation for a training batch with 3 samples, each with 2 rollouts. For each prompt-image pair $s_i$, the policy samples $G=2$ rollouts and we compute the task reward $R_T$, the consistency reward $R_C$, and grounding rewards $R_S$ and $R_G$ (only $R_G$ is shown for clarity). The consistency and semantic-grounding rewards are provided by an online VLM judge. Each signal is independently normalized within the rollout group to obtain decoupled advantages, which are then combined into the final FGRPO advantage using Lagrange multipliers. Batch-level constraint statistics are used to update the multipliers via dual ascent, increasing the weight of violated constraints and decreasing the weight of satisfied ones.}
    \label{fig:training_flow}
\end{figure*}

\subsection{Faithful GRPO (FGRPO)}
\label{subsec:fgrpo}

Rather than incorporating reasoning quality signals as reward terms where they compete with and are traded off against task accuracy, we propose treating them as \emph{constraints} that must be satisfied during policy optimization. This formulation ensures that the model cannot sacrifice reasoning quality for accuracy gains.

\paragraph{Constrained Optimization Formulation.}
Let $\pi_\theta$ denote the policy and $R_{\mathrm{task}}$ denote the task reward. We formulate FGRPO as a constrained optimization problem over the two reasoning quality axes: consistency and visual grounding.
\begin{equation}
\label{eq:fgrpo_objective}
\max_{\theta} \; \mathbb{E}_{x, o \sim \pi_\theta} \big[ R_{\mathrm{task}}(o) \big] \quad \text{s.t.} \quad \mathbb{E}[R_C(o)] \geq \tau_C, \quad \mathbb{E}[R_S(o)] \geq \tau_S, \quad \mathbb{E}[R_G(o)] \geq \tau_G,
\end{equation}
where $\tau_C$, $\tau_S$, and $\tau_G$ are threshold hyperparameters. The consistency constraint $R_C$ and semantic-grounding constraint $R_S$ operates on all correct-answer samples (Appendix~\ref{appx:semantic_grounding_details}). The spatial grounding constraint $R_G(o)$ is computed only for samples with bounding-box annotations.

\paragraph{Lagrangian Relaxation.}
Following the constrained MDP framework~\citep{altman1999constrained,tessler2019rcpo}, we convert the constrained problem into an unconstrained Lagrangian:
\begin{equation}
\label{eq:lagrangian}
\mathcal{L}(\theta, \boldsymbol{\lambda}) = \mathbb{E}\big[R_{\mathrm{task}}(o)\big] + \lambda_C \big(\mathbb{E}[R_C(o)] - \tau_C\big) + \lambda_S \big(\mathbb{E}[R_S(o)] - \tau_S\big) + \lambda_G \big(\mathbb{E}[R_G(o)] - \tau_G\big),
\end{equation}
where $\boldsymbol{\lambda} = (\lambda_C, \lambda_S, \lambda_G) \geq 0$ are Lagrange multipliers. The policy parameters $\theta$ are updated to maximize $\mathcal{L}$, while the multipliers are updated via dual ascent:
\begin{equation}
\label{eq:dual_ascent}
\lambda_k \leftarrow \mathrm{clip}\Big(0, \; \lambda_{\max}, \; \lambda_k + \eta_\lambda \cdot (\tau_k - \bar{c}_k)\Big), \quad k \in \{C, S, G\},
\end{equation}
where $\eta_\lambda$ is the dual learning rate and $\lambda_{\max}$ is an upper bound for stability. The batch-average constraint score $\bar{c}_k$ is computed over all $N \times G$ rollouts using per-rollout masks $m_k(o^j_i) \in \{0,1\}$:
\begin{equation}
\label{eq:batch_constraint}
\bar{c}_k = \frac{\textstyle\sum_{j=1}^{N}\sum_{i=1}^{G} m_k(o^j_i) \cdot R_k(o^j_i)}{\textstyle\sum_{j=1}^{N}\sum_{i=1}^{G} m_k(o^j_i)}, 
\end{equation}
where $m_C(o^j_i) = m_S(o^j_i) = \mathbb{1}[R_{\mathrm{acc}}(o^j_i) > 0]$ and $m_G(o^j_i) = \mathbb{1}[\text{source}(x_j) \in \{\text{VGR}, \text{TreeVGR}\}]$. Masking $R_C$ and $R_S$ to correct predictions prevents reward hacking: without masking, the model could trivially satisfy consistency and grounding constraints by producing incorrect but internally coherent and well-grounded answers, sacrificing accuracy to inflate constraint scores. Spatial grounding is masked by data availability since ground-truth bounding boxes are required. When $\bar{c}_k \geq \tau_k$, $\lambda_k$ decreases; when $\bar{c}_k < \tau_k$, $\lambda_k$ increases.

\paragraph{Practical Realization within GRPO.}
Training proceeds via alternating optimization. In the \emph{primal step}, the policy $\pi_\theta$ is updated to maximize the Lagrangian (Eq.~\ref{eq:lagrangian}) using the clipped surrogate objective from GRPO. In the \emph{dual step}, the multipliers are updated via Eq.~\ref{eq:dual_ascent} using the constraint scores from the current batch. Since GRPO operates on rollout-level rewards that are group-normalized into advantages, we must translate the Lagrangian into this advantage-based framework. The challenge is that the constraint signals are heterogeneous: $R_C$ is binary, $R_G$ is continuous and only defined for a subset of samples, and $R_S$ is a mean over per-sentence scores, so naively summing them into a single reward before normalization would allow one signal's scale to dominate. We address this via decoupled normalization.

\paragraph{Advantage Computation.}
A straightforward approach would be to add constraint rewards directly to $R_{\mathrm{task}}$ as weighted terms before group normalization. However, GRPO's within-group normalization can nullify such signals entirely: if a constraint score is constant across all rollouts for a prompt (e.g., all rollouts for $x_j$ receive the same binary consistency score), it cancels in the mean subtraction and contributes zero gradient, regardless of its weight. To avoid this, following GDPO~\citep{gdpo}, we apply group-relative normalization \emph{independently} to each signal before combining them. The final advantage for rollout $o^j_i$ is:
\begin{equation}
\label{eq:fgrpo_advantage}
\hat{A}_{\mathrm{FGRPO}}(o^j_i) = \hat{A}_{\mathrm{task}}(o^j_i) + \sum_{k \in \{C, S, G\}} \lambda_k \cdot \hat{A}_k(o^j_i),
\end{equation}
where $\hat{A}_{\mathrm{task}}(o^j_i)$ and $\hat{A}_k(o^j_i)$ are independently group-normalized advantages for the task reward and constraint $k$ within group $j$, as shown in Figure~\ref{fig:training_flow}. This decoupled normalization ensures each signal operates on a comparable scale and contributes meaningful gradient even when its within-group variance is low. The combined advantage is whitened and used in the standard clipped surrogate loss. Note that decoupled normalization with \emph{fixed} multipliers already resolves signal cancellation, but converges to a single point on the Pareto frontier dictated by the weight vector, with no mechanism to enforce the thresholds $\tau_k$. Adaptive dual ascent (Eq.~\ref{eq:dual_ascent}) addresses this by growing $\lambda_k$ when a constraint is violated and shrinking it when satisfied, producing an emergent curriculum that automatically prioritizes whichever constraint is most violated without manual weight tuning (Table~\ref{tab:ablation}).

\label{subsec:semantic_grounding_reward}

\section{Experiments}
\label{sec:experiments}

We evaluate FGRPO on two model scales across seven spatial reasoning benchmarks, comparing against unconstrained GRPO baselines and reward design alternatives. We then ablate the contribution of individual constraints, their composition, and the role of adaptive Lagrange multipliers.

\subsection{Experimental Setup}
\label{subsec:experimental_setup}

\textbf{Models.}
We train two backbone sizes: Qwen2.5-VL-3B-Instruct and Qwen2.5-VL-7B-Instruct~\citep{qwen2.5vl_techreport}, both following the two-stage pipeline. We curate bounding-box grounded CoT data, where each reasoning step references regions of interest via \texttt{<bbox>[x1,y1,x2,y2]</bbox>} tags before reasoning over salient objects. The SFT checkpoint trained on this bbox CoT data serves as the common initialization for all RL models.

\textbf{Training Data.}
The RL training set comprises approximately 36K samples drawn from SAT~\citep{ray2025sat}, VGR~\citep{wang2025vgrvisualgroundedreasoning}, and VisCoT~\citep{shao2024visual}, supplemented with 13K samples from TreeVGR-RL-37K~\citep{treevgr} for bounding-box supervision diversity. Samples are filtered by intermediate difficulty following prior work~\citep{less,Yu2025DAPOAO}. Ground-truth bounding-box annotations are available only for VGR and TreeVGR; the spatial grounding constraint is masked accordingly (\S\ref{subsec:fgrpo}).

\textbf{Evaluation.}
We evaluate on seven spatial reasoning benchmarks: CVBench-2D~\citep{cambrian1}, CVBench-3D~\citep{cambrian1}, MindCube~\citep{mindcube}, MMVP~\citep{MMVP}, OmniSpatial~\citep{omnispatial25}, RealWorldQA~\citep{RealWorldQA_xai}, and SAT-Real~\citep{ray2025sat}, spanning both in-distribution and out-of-distribution settings. We report three metrics: (i)~\textbf{Accuracy} (pass@1 with greedy decoding), (ii)~\textbf{Inconsistency Rate} ($\mathrm{IR} = N_{\text{inconsistent}} / N_{\text{total}}$), the fraction of all samples whose reasoning trace is inconsistent with the final answer (lower is better), and (iii)~\textbf{Semantic Grounding} $R_S(o)$, the mean per-sentence visual grounding score from the VLM judge (\S\ref{subsec:diagnosing}). During training, consistency and semantic grounding rewards are computed using Qwen3-VL-30B-A3B-Instruct as the online judge. At evaluation time, we use GPT-5.4~\citep{openai2025gpt5} as the judge, ensuring that evaluation is independent of the training reward model. We validate judge reliability in Appendix~\ref{appx:training_details}. 

\textbf{Hyperparameters.}
All RL runs use AdamW with learning rate $1 \times 10^{-6}$, bf16 precision, $8\times$H100 GPUs, $G = 5$ rollouts per prompt, and a KL coefficient of $0.001$. For FGRPO, the Lagrange dual learning rate is $\eta_\lambda = 0.05$, the multiplier upper bound is $\lambda_{\max} = 5.0$, and constraint updates require a minimum of 8 applicable samples per batch. Unless otherwise noted, constraint thresholds are $\tau_C = 0.95$, $\tau_G = 0.7$, $\tau_S = 0.95$, and initial multipliers are $\lambda_C^{(0)} = \lambda_G^{(0)} = \lambda_S^{(0)} = 1.0$. Full configuration details are in Appendix~\ref{appx:training_details}.

\begin{table*}[t]
\centering
\caption{\textbf{Main results.} Pass@1 accuracy on seven spatial reasoning benchmarks. All 7B models use Qwen2.5-VL-7B-Instruct as backbone; 3B models use Qwen2.5-VL-3B-Instruct. \colorbox{firstbg}{\textbf{Bold}} = best, \colorbox{secondbg}{\underline{underline}} = second best among open-weight 7B models. FGRPO achieves the highest average accuracy at both scales, outperforming all MRM baselines and the unconstrained GRPO-T baseline while simultaneously improving reasoning quality.}
\label{tab:main_results}
\vspace{0.5em}
\small
\resizebox{\textwidth}{!}{
\begin{tabular}{@{}l|ccccccc|c@{}}
\toprule
\textbf{Method} & \makecell{\textbf{CVB}\\\textbf{2D}} & \makecell{\textbf{CVB}\\\textbf{3D}} & \makecell{\textbf{Mind}\\\textbf{Cube}} & \textbf{MMVP} & \makecell{\textbf{Omni}\\\textbf{Spatial}} & \makecell{\textbf{Real}\\\textbf{WorldQA}} & \makecell{\textbf{SAT}\\\textbf{Real}} & \textbf{Avg.} \\
\midrule
\multicolumn{9}{@{}l}{\emph{Base model (7B): Qwen2.5-VL-7B-Instruct}} \\
\quad Non-reasoning          & 77.17 & 83.78 & 35.11 & 75.78 & 45.23 & 69.02 & 63.11 & 64.17 \\
\quad CoT prompting          & 75.92 & 76.09 & 30.83 & 72.44 & 40.40 & 63.05 & 59.22 & 59.71 \\
\multicolumn{9}{@{}l}{\emph{Base model (3B): Qwen2.5-VL-3B-Instruct}} \\
\quad Non-reasoning          & 70.58 & 74.00 & 43.71 & 64.67 & 45.92 & 65.88 & 59.00 & 60.54 \\
\quad CoT prompting          & 71.21 & 68.00 & 40.38 & 63.67 & 40.77 & 62.35 & 55.00 & 57.34 \\
\midrule
\multicolumn{9}{@{}l}{\emph{MRM baselines}} \\
\quad R1-OneVision~\citep{r1onevision}   & 53.31 & 58.00 & 27.09 & 56.16 & 31.54 & 49.87 & 51.50 & 46.78 \\
\quad Vision-R1~\citep{visionr1}         & 71.58 & 75.83 & 36.95 & 72.22 & 39.75 & 67.41 & 58.45 & 60.31 \\
\quad ViGoRL-Spatial~\citep{vigorl}      & 76.59 & \cellcolor{secondbg}\underline{86.14} & 39.36 & 73.22 & 36.97 & 65.67 & 58.44 & 62.34 \\
\quad TreeVGR~\citep{treevgr}            & 74.69 & 73.92 & \cellcolor{secondbg}\underline{43.14} & 71.00 & \cellcolor{firstbg}\textbf{45.99} & 66.80 & 61.00 & 62.36 \\
\quad VL-Rethinker~\citep{vlrethinker}   & 76.06 & 80.75 & 37.81 & \cellcolor{firstbg}\textbf{75.89} & 39.84 & \cellcolor{firstbg}\textbf{68.50} & \cellcolor{secondbg}\underline{65.00} & \cellcolor{secondbg}\underline{63.41} \\
\midrule
\multicolumn{9}{@{}l}{\emph{Proprietary models}} \\
\quad GPT-5-nano (CoT)                    & 76.29 & 86.75 & 27.71 & 75.67 & 41.03 & 71.90 & 64.00 & 63.34 \\
\quad GPT-4o (CoT)          & 78.23 & 86.42 & 43.52 & 84.33 & 45.73 & 73.59 & 68.67 & 68.64 \\
\midrule
\multicolumn{9}{@{}l}{\emph{Ours (7B)}} \\
\quad GRPO-T                          & \cellcolor{secondbg}\underline{79.97} & 85.92 & 41.71 & \cellcolor{secondbg}\underline{74.00} & 40.90 & 66.67 & \cellcolor{firstbg}\textbf{67.00} & \cellcolor{secondbg}\underline{65.17} \\
\quad \textbf{FGRPO} & \cellcolor{firstbg}\textbf{82.38} & \cellcolor{firstbg}\textbf{87.04} & \cellcolor{firstbg}\textbf{49.28} & 73.33 & \cellcolor{secondbg}\underline{44.78} & \cellcolor{secondbg}\underline{67.64} & 65.66 & \cellcolor{firstbg}\textbf{67.16} \\
\midrule
\multicolumn{9}{@{}l}{\emph{Ours (3B)}} \\
\quad GRPO-T                          & 77.24 & 80.44 & 46.06 & 65.22 & 39.88 & 61.83 & 58.67 & 61.33 \\
\quad \textbf{FGRPO} & 77.95 & 82.5 & 49.1 & 64.3 & 43.6 & 60.7 & 58.6 & 62.39 \\
\bottomrule
\end{tabular}
}
\end{table*}

\subsection{Main Results}
\label{subsec:main_results}

Table~\ref{tab:main_results} compares FGRPO against the Qwen2.5-VL backbone, five contemporary MRMs, and GRPO-T (task reward only). We make three observations:
First, the two-stage pipeline with diverse data curation yields a strong baseline: GRPO-T alone outperforms most existing MRM baselines, including TreeVGR and ViGoRL-Spatial, which are explicitly trained for visual spatial reasoning.
Second, FGRPO further improves upon this strong baseline, lifting average accuracy from 65.17 to 67.16 with consistent gains across the majority of benchmarks.
Third, the same pattern holds at the 3B scale (Table~\ref{tab:main_results}, bottom), confirming that FGRPO's gains are not specific to the 7B backbone.

\begin{figure*}[t]
\centering
\includegraphics[width=\textwidth]{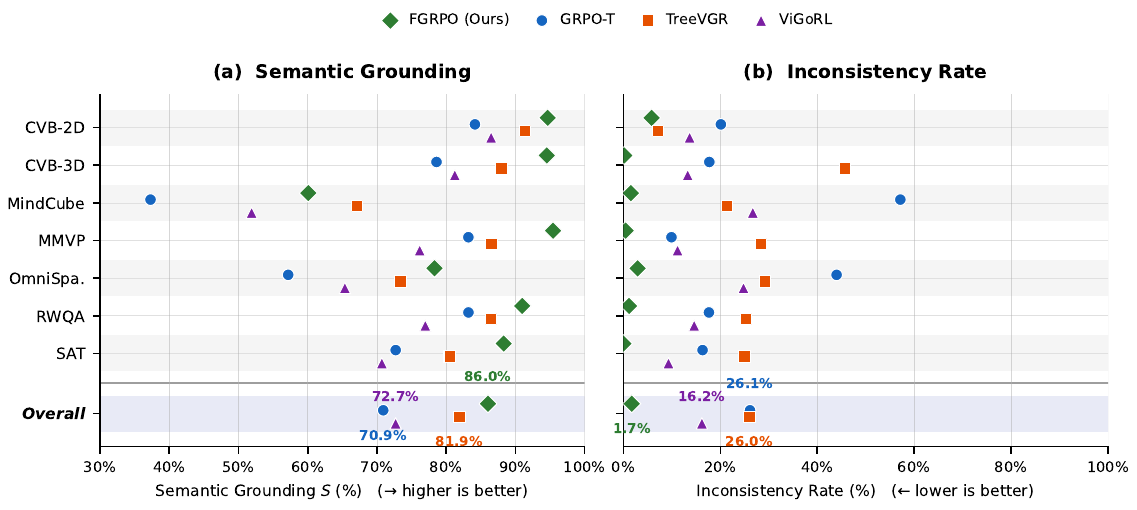}
\caption{\textbf{Per-dataset reasoning quality breakdown}. \textbf{(a)}~Semantic grounding~($S$): FGRPO achieves uniformly higher semantic grounding than GRPO-T across all seven benchmarks (86.0\% vs.\ 72.7\%), with the largest gains on MindCube (+22.8\,pp) and OmniSpatial (+21.1\,pp). TreeVGR also outperforms GRPO-T (81.9\%). \textbf{(b)}~Inconsistency rate: FGRPO reduces inconsistency to 1.7\% on average, compared to 26.1\% (GRPO-T), 26.0\% (TreeVGR), and 16.2\% (ViGoRL). MindCube and OmniSpatial are the most challenging for all baselines; FGRPO virtually eliminates inconsistency on every benchmark.}
\label{fig:faithfulness_consistency}
\end{figure*}

\subsection{Reasoning Quality}
\label{subsec:reasoning_quality}
\textbf{Inconsistency \& Semantic Grounding.} Figure~\ref{fig:faithfulness_consistency} breaks down reasoning quality by dataset. FGRPO achieves higher semantic grounding than GRPO-T across all seven benchmarks (86.0\% vs.\ 72.7\% overall), with the largest gains on MindCube (+22.8\,pp) and OmniSpatial (+21.1\%). These datasets require multi-step spatial reasoning where unconstrained models produce the most ungrounded sentences. On MindCube, FGRPO nearly doubles the grounding rate of GRPO-T (60.1\% vs.\ 37.3\%). TreeVGR also outperforms GRPO-T on semantic grounding (81.9\%), but still trails FGRPO by 4.1\%. FGRPO reduces inconsistency to 1.7\% on average, compared to 26.1\% (GRPO-T), 26.0\% (TreeVGR), and 16.2\% (ViGoRL). Inconsistency is most acute on MindCube (57.1\% for GRPO-T) and OmniSpatial (44.0\%) but FGRPO virtually eliminates it, achieving near-zero inconsistency on six of seven benchmarks.

\subsection{Accuracy vs Consistency Tradeoff in RLVR}
\label{subsec:pareto}
\begin{figure*}[t]
\centering
\includegraphics[width=\textwidth]{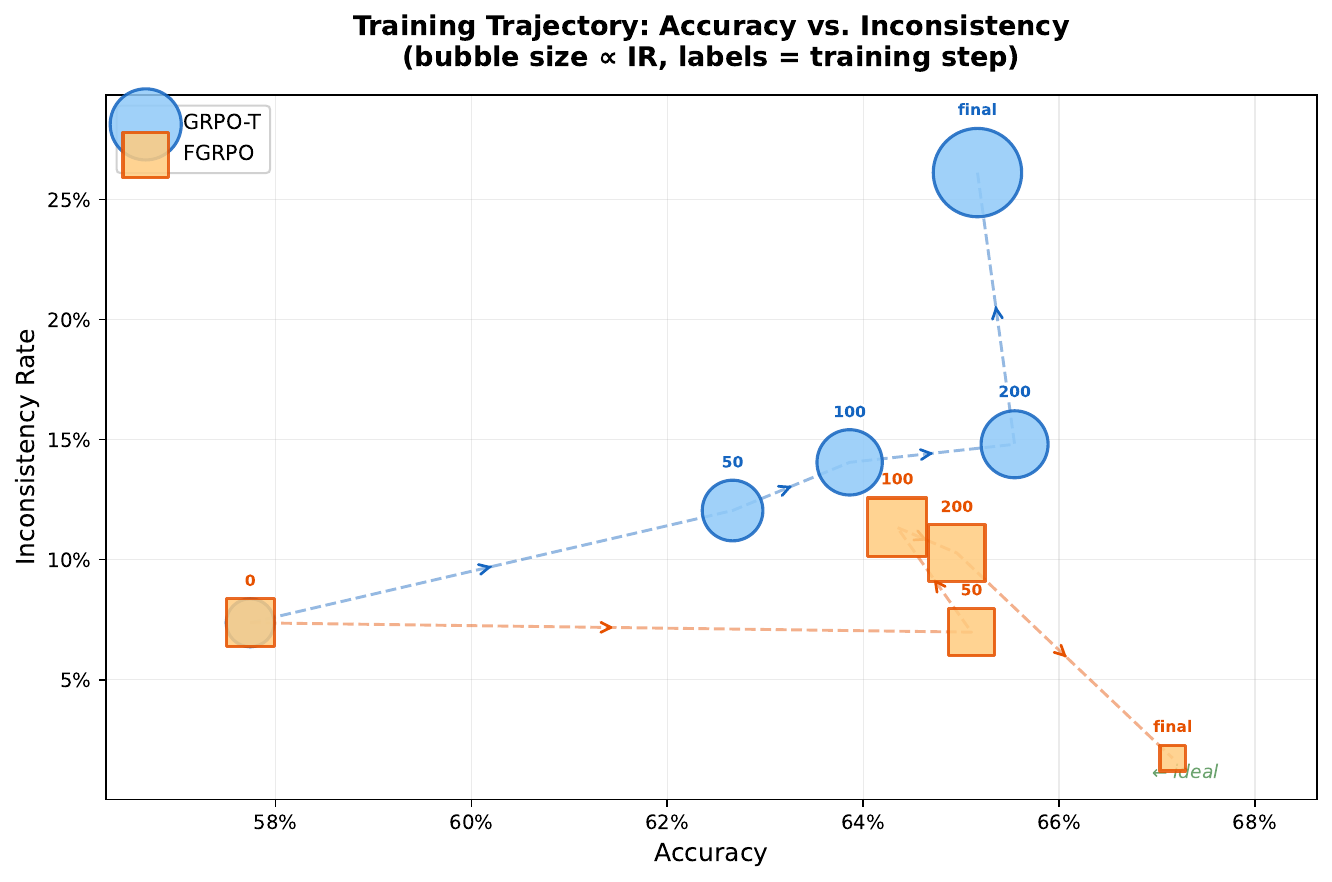}
\vspace{-4pt}
\caption{\footnotesize FGRPO (squares) vs GRPO-T (circles); larger shape size indicates higher inconsistency.}
\label{fig:pareto}
\end{figure*}
Figure~\ref{fig:pareto} traces the accuracy–inconsistency front across training iterations for GRPO-T and FGRPO, starting from an identical warmstart checkpoint. For GRPO-T accuracy climbs from 57.7\% to 65.1\% over 1000 steps, but inconsistency more than triples (7.4\%→26.1\%), and it produces increasingly unfaithful reasoning chains even as answer correctness improves. FGRPO on the other hand keeps inconsistency near or below the starting rate throughout training while pushing accuracy to 67.2\%, converging at 1.7\% inconsistency. 

\subsection{Reward Design and Multiplier Strategy}
\label{subsec:ablation}

\begin{wraptable}[9]{r}{0.5\textwidth}
\vspace{-1cm}
\centering
\caption{First three rows are GRPO.}
\label{tab:reward_shaping}
\small
\scalebox{0.9}{
\begin{tabular}{@{}lcc@{}}
\toprule
\textbf{Method} & \textbf{Acc (\%)} $\uparrow$ & \textbf{IR (\%)} $\downarrow$ \\
\midrule
\multicolumn{3}{@{}l}{\emph{GRPO (coupled advantage)}} \\
\quad Task only (GRPO-T)               & 65.16 & 26.12 \\
\quad + Additive $R_C$, $R_G$          & 64.97 &  4.22 \\
\quad + Multiplicative $R_C$, $R_G$    & 63.47 & 19.61 \\
\midrule
\multicolumn{3}{@{}l}{\emph{FGRPO (decoupled advantage)}} \\
\quad Consistency only ($R_C$)         & 66.16 & \textbf{0.54} \\
\quad \textbf{Full} ($R_C + R_S + R_G$) & \textbf{67.16} &  1.73 \\
\bottomrule
\end{tabular}
}    
\end{wraptable}

\textbf{Coupled vs.\ decoupled advantage.}
The top block of Table~\ref{tab:reward_shaping} shows three GRPO variants that incorporate $R_C$ and $R_G$ as additional reward terms, combined into a single scalar before group normalization. Adding consistency and grounding additively ($R = \tfrac{1}{3}R_{\mathrm{acc}} + \tfrac{1}{3}(R_{\mathrm{acc}} \cdot R_C) + \tfrac{1}{3}(R_{\mathrm{fmt}} \cdot R_G)$) reduces inconsistency to 4.2\% but sacrifices accuracy ($-0.2$). Multiplicative gating ($R = 0.5 \cdot R_{\mathrm{acc}} \cdot R_C + 0.5 \cdot R_{\mathrm{fmt}} \cdot R_G$) fares worse: accuracy drops by 1.7 points while inconsistency remains at 19.6\%. The bottom block uses FGRPO's decoupled advantage formulation (Eq.~\ref{eq:fgrpo_advantage}), where each signal is independently normalized. Even with only the consistency constraint, FGRPO improves both accuracy (+1.0) and inconsistency (0.54\%), confirming that decoupled normalization resolves the signal cancellation problem identified in \S\ref{subsec:fgrpo}. Adding the grounding constraints ($R_S$ and $R_G$) further lifts accuracy to 67.16 (+2.0 over GRPO-T).


\begin{wraptable}[6]{r}{0.5\textwidth}
\vspace{-0.8cm}
\centering
\caption{Adaptive $\lambda$s yield best accuracy}
\label{tab:ablation}
\small
\scalebox{0.9}{
\begin{tabular}{@{}lcc@{}}
\toprule
\textbf{Method} & \textbf{Acc (\%)} $\uparrow$ & \textbf{IR (\%)} $\downarrow$ \\
\midrule
GRPO-T                        & 65.17 & 26.12 \\
FGRPO (fixed $\lambda$)          & 66.32 &  1.11 \\
\textbf{FGRPO} (adaptive $\lambda$)  & \textbf{67.16} &  1.73 \\
\bottomrule
\end{tabular}
}    
\end{wraptable}
\textbf{Adaptive vs.\ fixed multipliers.} Table~\ref{tab:ablation} compares adaptive dual ascent against fixed multipliers ($\lambda_k = 1.0$, no updates). Fixed multipliers already yield strong consistency (IR 1.11\%) and a +1.2 accuracy gain, confirming that the decoupled advantage formulation is the primary driver. Adaptive dual ascent (Eq.~\ref{eq:dual_ascent}) further improves accuracy to 67.16 by reallocating optimization pressure as constraints are progressively satisfied.

\section{Conclusion}
\label{sec:conclusion}

We presented Faithful GRPO (FGRPO), a constrained variant of GRPO that enforces logical consistency and visual grounding as hard constraints during policy optimization for multimodal reasoning. By defining verifiable reward signals for consistency (via an LLM judge) and grounding (via per-sentence VLM scoring and CIoU-based spatial matching), and enforcing them through Lagrangian dual ascent with decoupled normalization, FGRPO ensures that accuracy gains do not come at the expense of reasoning quality. On seven spatial reasoning benchmarks, FGRPO reduces the inconsistency rate from 26.1\% to 1.7\%, improves semantic grounding by +13 percentage points, and simultaneously raises answer accuracy by +2\% over standard GRPO demonstrating that faithful reasoning and accurate answers are complementary objectives. We hope that FGRPO encourages the community to move beyond accuracy-only evaluation and to treat reasoning quality as a first-class objective in multimodal RL training.




\bibliography{colm2026_conference}
\bibliographystyle{colm2026_conference}

\newpage
\appendix
\section*{Appendix: Faithful GRPO: Improving Visual Spatial Reasoning in Multimodal Language Models via Constrained Policy Optimization}

\section{Training and Data Curation Details}
\label{appx:training_details}

In this appendix we provide comprehensive details on the training pipeline, data curation, evaluation setup, and reward computation summarized in the main paper.

\subsection{Semantic Grounding Reward: Sentence Decomposition and VLM Judge}
\label{appx:semantic_grounding_details}

The semantic grounding reward $R_S(o)$ requires evaluating whether each reasoning step in the CoT trace accurately reflects the image content. We decompose the reasoning trace into individual sentences and score each via a VLM judge.

\paragraph{Sentence Decomposition.}
Given a response $o$ with reasoning trace within \texttt{<think>} tags, we decompose the trace into individual sentences $\{s_1, s_2, \ldots, s_K\}$. Each sentence is classified as either \emph{visual} (makes a specific claim about image content) or \emph{trivial} (meta-reasoning, planning, hedging, arithmetic). Trivial sentences are identified via pattern matching against common prefixes (e.g., ``Let me...'', ``Therefore...'', ``Wait...'') and the absence of visual keywords (e.g., object names, colors, spatial terms, coordinate references). Trivial sentences are excluded from VLM evaluation to reduce computational cost and avoid penalizing non-visual reasoning steps.

\paragraph{VLM-as-Judge Scoring.}
Each non-trivial sentence $s_k$ is evaluated by a VLM judge that receives the image, the question, and the reasoning context (all preceding sentences). The judge classifies the sentence into one of three categories:
\begin{itemize}[nosep,leftmargin=*]
    \item \textbf{CORRECT} ($\hat{s}_k = 1$): The sentence makes a specific visual claim that is accurate---objects are present, attributes match, spatial relationships are correct, and any referenced bounding boxes contain the described content.
    \item \textbf{INCORRECT} ($\hat{s}_k = 0$): The sentence makes a specific visual claim that is inaccurate, or repeats a previously stated claim without adding new visual evidence.
    \item \textbf{SKIP}: The sentence makes no specific visual claim. These are excluded from the reward computation entirely, similar to the trivial sentence filter.
\end{itemize}
\noindent The full judge prompt is provided in Appendix~\ref{appx:prompts}. We validate judge reliability against human annotations in Appendix~\ref{appx:judge_validation}.

\subsection{Two-Stage Training Pipeline}
\label{appx:two_stage_pipeline}

\begin{figure*}[h]
    \centering
    \includegraphics[width=\linewidth]{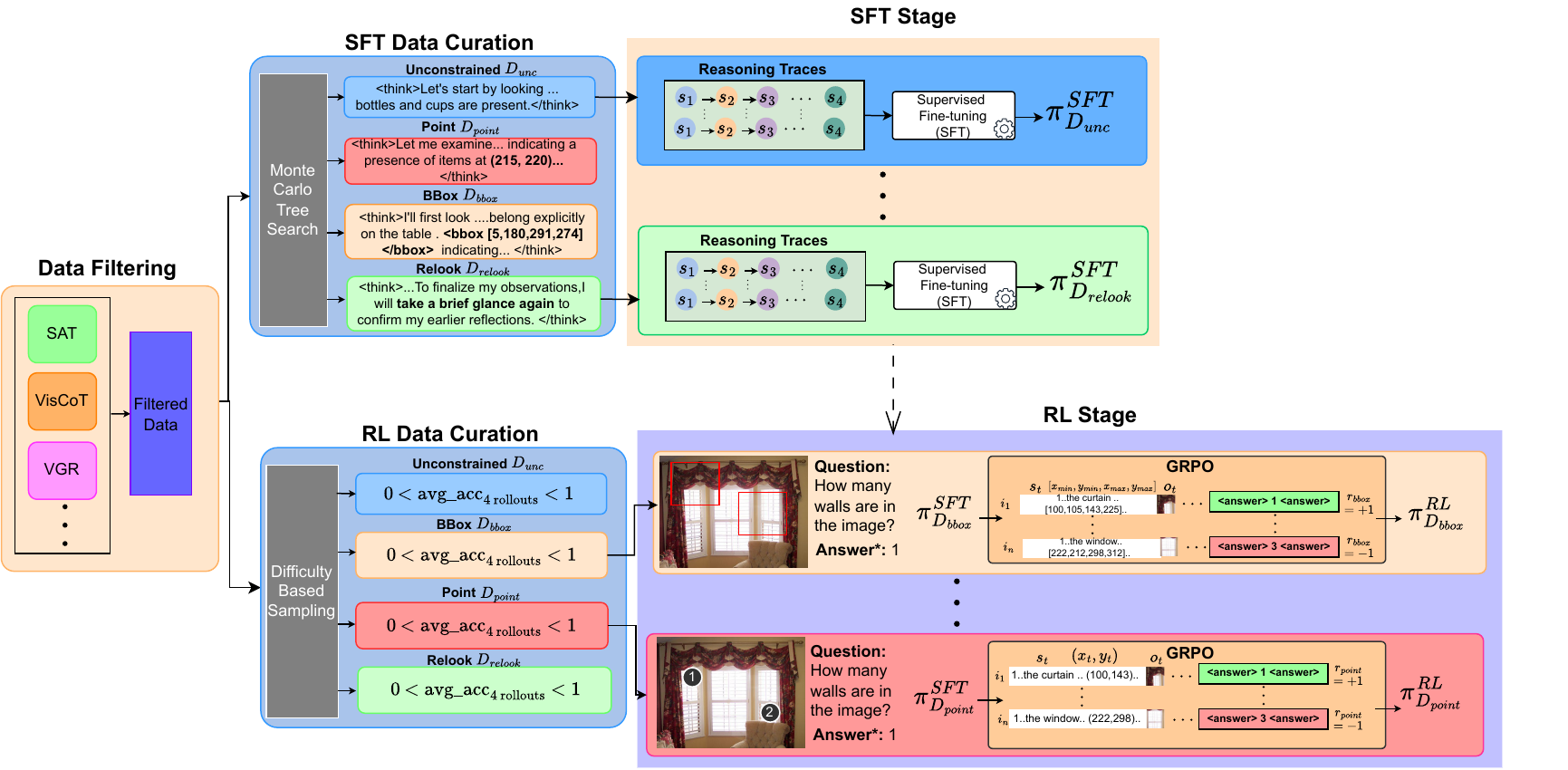}
    \caption{An overview of the two-stage training pipeline. We curate CoT training data using MCTS with a strong visual teacher, covering bounding-box grounded visual reasoning. We perform SFT on the curated CoT data using data from SAT~\citep{ray2025sat}, VGR~\citep{wang2025vgrvisualgroundedreasoning} and VisCoT~\citep{shao2024visual}, followed by RL training. We add a subset of TreeVGR-RL-37K~\citep{treevgr} to construct diverse RL data with bounding-box supervision.}
    \label{fig:two_stage_pipeline}
\end{figure*}

We follow the two-stage training paradigm of DeepSeek-R1~\citep{DeepSeekAI2025DeepSeekR1IR}. In stage one, we perform supervised finetuning (SFT) on curated Chain-of-Thought (CoT) data to imbue the model with baseline spatial reasoning capability. The MLM policy $\pi_{\theta}(o|x)$ autoregressively generates text $o=(s_1, s_2,\dots, s_T, a)$ conditioned on a multimodal prompt $x$, where $s_t$ denotes a reasoning step and $a$ denotes the final answer. For each SFT training sample $(x, o=(o_1,o_2,\dots,o_T))$, the negative log-likelihood $L_{\mathrm{sft}}=-\mathbb{E}_{(x,o)\sim D}\sum_{t=1}^T \log\pi_{\theta}(o_t|x,o_{<t})$ is minimized. We use LLaMA-Factory~\citep{zheng2024llamafactory} for SFT with $3$ epochs, AdamW~\citep{adamw} optimizer (learning rate $1\mathrm{e}{-6}$, weight decay $0.01$), batch size $32$, and cosine learning rate schedule with warmup ratio $0.03$. We freeze the vision encoder, keeping the LLM backbone and the visual projector trainable. The best checkpoint is selected based on validation loss. SFT is performed on $4\times$ NVIDIA A100 80GB GPUs using DeepSpeed ZeRO Stage 3 and takes approximately 12 hours per variant.

In stage two, the SFT checkpoint is finetuned using RLVR with Group Relative Policy Optimization (GRPO)~\citep{grpo} to learn generalizable reasoning behaviors beyond the training distribution. We use a modified version of Easy-R1~\citep{easyr1} for RL training. RL training uses GRPO with a group size of $G=5$ rollouts per prompt, learning rate $1\mathrm{e}{-6}$, and rollout batch size of $128$. vLLM~\citep{vllm} is used to generate rollouts efficiently. RL training is performed on $8\times$ NVIDIA H100 GPUs. The full hyperparameter configurations for both stages are provided in Tables~\ref{tab:sft_hyperparams} and~\ref{tab:grpo_hyperparams}.

\begin{table}[h]
    \centering
    \small
    \begin{minipage}[t]{0.48\textwidth}
        \centering
        \caption{SFT Hyperparameters}
        \label{tab:sft_hyperparams}
        \begin{tabular}{lc}
            \toprule
            \textbf{Hyperparameter} & \textbf{Value} \\
            \midrule
            Epochs & 3 \\
            Learning rate & 1e-6 \\
            Weight decay & 0.01 \\
            Warmup ratio & 0.03 \\
            Batch size (per device) & 4 \\
            Gradient accumulation & 2 \\
            Scheduler & Cosine \\
            Precision & bf16 \\
            Flash attention & fa2 \\
            Freeze vision tower & True \\
            Max sequence length & 8192 \\
            Deepspeed config & ZeRO Stage 3 \\
            \bottomrule
        \end{tabular}
    \end{minipage}
    \hfill
    \begin{minipage}[t]{0.48\textwidth}
        \centering
        \caption{GRPO Training Hyperparameters}
        \label{tab:grpo_hyperparams}
        \begin{tabular}{lc}
            \toprule
            \textbf{Hyperparameter} & \textbf{Value} \\
            \midrule
            Total Epochs & 3 \\
            Learning rate & 1e-6 \\
            Weight decay & 0.01 \\
            Warmup ratio & 0.0 \\
            Optimizer & AdamW (bf16) \\
            Group size ($G$) & 5 \\
            KL coefficient & 0.01 \\
            Clip ratio & 0.28 \\
            Gradient clipping & 1.0 \\
            Rollout batch size & 128 \\
            Global batch size & 64 \\
            Rollout engine & vLLM \\
            Max prompt length & 8192 \\
            Max response length & 1024 \\
            Freeze vision tower & True \\
            Precision & bf16 \\
            \bottomrule
        \end{tabular}
    \end{minipage}
\end{table}

For FGRPO training, we additionally specify constraint-specific hyperparameters in Table~\ref{tab:fgrpo_hyperparams}.

\begin{table}[h]
    \centering
    \small
    \caption{FGRPO Constraint Hyperparameters}
    \label{tab:fgrpo_hyperparams}
    \begin{tabular}{lc}
        \toprule
        \textbf{Hyperparameter} & \textbf{Value} \\
        \midrule
        Advantage estimator & CGRPO \\
        Constraint keys & [consistency, bbox\_reward, faithfulness] \\
        \midrule
        \emph{Consistency constraint} \\
        \quad Threshold ($\tau_C$) & 0.95 \\
        \quad Initial $\lambda_C$ & 1.0 \\
        \quad Masking & Only correct predictions ($R_{\mathrm{acc}} > 0$) \\
        \midrule
        \emph{Spatial grounding constraint} \\
        \quad Threshold ($\tau_G$) & 0.65 \\
        \quad Initial $\lambda_G$ & 1.0 \\
        \quad Masking & Only VGR/TreeVGR samples \\
        \midrule
        \emph{Semantic grounding constraint} \\
        \quad Threshold ($\tau_S$) & 0.95 \\
        \quad Initial $\lambda_S$ & 1.0 \\
        \quad Reward level & Outcome (with per-sentence rewards) \\
        \quad Masking & Only correct predictions ($R_{\mathrm{acc}} > 0$) \\
        \midrule
        Dual learning rate ($\eta_\lambda$) & 0.05 \\
        Max Lagrange multiplier ($\lambda_{\max}$) & 5.0 \\
        Min applicable samples & 8 \\
        \bottomrule
    \end{tabular}
\end{table}

\subsection{Training Data Curation}
\label{appx:data_curation}

We curate both SFT and RL datasets from three seed sources that span diverse real-world spatial questions and images.

\paragraph{Seed Datasets.}
We select SAT~\citep{ray2025sat} (150K samples), VGR~\citep{wang2025vgrvisualgroundedreasoning} (90K samples), and VisCoT~\citep{shao2024visual} (363K samples) as seed datasets. These datasets span common image sources such as COCO~\citep{lin2015microsoftcococommonobjects}, GQA~\citep{hudson2019gqa}, OpenImages~\citep{openimages}, and Flickr30k~\citep{young2014image}, and cover a diverse range of spatial questions. We discard irrelevant data such as charts, tables, and visual math questions.

\paragraph{CoT Data Curation for SFT.}
We curate CoT data by distilling reasoning chains from a strong visual teacher (Qwen2.5-VL-72B-Instruct). For each seed domain, we cluster question embeddings and pick representative samples from each cluster: 1.5K from SAT, 1.5K from VGR, and 3K from VisCoT, for a total of 6K seed samples. For each sample, we run Monte Carlo Tree Search (MCTS) to generate reasoning traces (see \S\ref{appx:mcts_procedure}). MCTS-based trajectories are not restricted to the base policy of the teacher, allowing us to generate approximately 45K high-quality diverse rollouts from the relatively small pool of 6K samples. We generate CoT data using the bounding-box grounded reasoning format, which produces reasoning chains with explicit \texttt{<bbox>} coordinate references.

\paragraph{RL Data Curation.}
We employ a difficulty-based data filtering strategy~\citep{less} on each seed dataset. For each sample, we generate 4 rollouts using the base model Qwen2.5-VL-7B-Instruct and compute the average accuracy. We preferentially select samples of intermediate difficulty (average accuracy neither 0 nor 1), since samples that are too easy or too hard do not contribute effectively to the GRPO training objective~\citep{Yu2025DAPOAO}. We include only 10\% of trivially easy or hard samples. We select 15.7K samples from SAT, 7K from VGR, and 13K from VisCoT. To further improve diversity, we add 13K samples from the TreeVGR-RL-37K~\citep{treevgr} dataset sourced from V* and VisDrone, for a final RL dataset of approximately 49K samples. TreeVGR-RL and VGR provide bounding-box annotations, which are required for the spatial grounding constraint. Unlike in SFT data curation, we construct a single RL dataset used by all training variants.

\subsection{MCTS Procedure}
\label{appx:mcts_procedure}

To synthesize high-quality spatial reasoning data, we formulate the reasoning process as a search problem over a decision tree $\mathcal{T}$, where the root node represents the initial multimodal input tuple $(I, Q)$. We utilize the Qwen2.5-VL-72B-Instruct model as the teacher policy $\pi_\theta$ to navigate this search space.

We define a state $s_t$ as the sequence of reasoning steps generated up to time step $t$, and an action $a_t$ as the generation of the next reasoning step. For the bounding-box grounded variant used in FGRPO, actions take the form $a_t \in \text{Text} \times [x_1, y_1, x_2, y_2]$, employing a progressive zoom strategy where the model starts with large regions for global context and progressively focuses on smaller, more specific subregions.

\paragraph{Search Algorithm.}
We employ the Predictor + Upper Confidence Bound (PUCT) algorithm to traverse the tree. Each iteration proceeds through four phases. During \textit{Selection}, the algorithm recursively selects child nodes starting from the root that maximize the PUCT objective:
\begin{equation}
    \text{PUCT}(s, a) = Q(s, a) + c_{\mathrm{puct}} \cdot P(s, a) \cdot \frac{\sqrt{\sum_b N(s, b)}}{1 + N(s, a)}
\end{equation}
where $Q(s, a)$ is the estimated value, $P(s, a)$ is the prior probability, and $N(s, a)$ is the visit count. We set $c_{\mathrm{puct}} = 2.0$ to promote exploration of diverse visual regions. Upon reaching a leaf node, \textit{Expansion} is performed if the current tree depth $d < d_{\max}$, prompting the teacher model to sample $k=3$ distinct reasoning steps. From each newly expanded node, \textit{Simulation} performs $n=2$ rollouts using high temperature ($T=1.0$) for diversity. Finally, in \textit{Backpropagation}, the terminal answer is scored via exact string matching against ground truth and the resulting binary reward is propagated to update ancestor node statistics.

\paragraph{Trajectory Linearization and Synthetic Backtracking.}
Post-search, we linearize the tree into SFT training data. Following~\citet{vigorl}, we extract two categories of traces: (i)~\textit{Direct Chains}---optimal paths from root to a correct terminal node with the highest reward, and (ii)~\textit{Corrected Chains}---trajectories that include an incorrect branch, a fixed backtracking cue (``\textit{Wait, this seems off. Let's try something else.}''), followed by the correct branch. We select approximately 10 shortest chains per sample (8 direct, 2 corrected), ensuring the student model learns both correct reasoning and self-correction. The MCTS hyperparameters are summarized in Table~\ref{tab:mcts_hyperparams}.

\begin{table}[h]
    \centering
    \small
    \caption{Hyperparameters for MCTS-based data synthesis.}
    \label{tab:mcts_hyperparams}
    \begin{tabular}{l|c}
        \toprule
        \textbf{Hyperparameter} & \textbf{Value} \\
        \midrule
        Teacher Model & Qwen2.5-VL-72B-Instruct \\
        $N_{\mathrm{sim}}$ & 8 \\
        $c_{\mathrm{puct}}$ & 2.0 \\
        Branching Factor ($k$) & 3 \\
        Rollouts per Node ($n$) & 2 \\
        Max Depth & 10 \\
        Judge & String Match \\
        \bottomrule
    \end{tabular}
\end{table}

\section{Evaluation Details}
\label{appx:eval_details}

\subsection{Evaluation Datasets and Protocol}
\label{appx:eval_datasets}

We evaluate on seven curated datasets that measure various facets of visual spatial reasoning, totaling approximately 6.3K challenging questions. Table~\ref{tab:appx_eval_datasets} summarizes the datasets.

\begin{table}[h]
    \centering
    \small
    \begin{tabular}{l|cc}
    \toprule
    Dataset & \#Questions & Subtasks \\
    \midrule
    CV-Bench2D~\citep{cambrian1} & 1.4K & 2D, spatial relations \\
    CV-Bench3D~\citep{cambrian1} & 1.2K & 3D, depth \\
    MindCube~\citep{mindcube} & 1K & 3D, multi-image \\
    MMVP~\citep{MMVP} & 300 & 2D, spatial relations \\
    OmniSpatial~\citep{omnispatial25} & 1.5K & 3D, dynamics \\
    RealWorldQA~\citep{RealWorldQA_xai} & 765 & 2D, real-world \\
    SAT-Real~\citep{ray2025sat} & 150* & 3D, interaction \\
    \bottomrule
    \end{tabular}
    \caption{Summary of evaluation datasets. Star (*) indicates circular evaluation is employed.}
    \label{tab:appx_eval_datasets}
\end{table}

We perform inference using vLLM~\citep{vllm} v0.11.0 on 4 NVIDIA A100 GPUs with batch size 16, max new tokens 4096, model context length 32768, and float16 precision. We keep original image resolution whenever the number of input tokens does not exceed 32768. We report pass@1 accuracy with greedy decoding. All models including ours are asked to generate reasoning within \texttt{<think>} tags before generating the final answer within \texttt{<answer>} tags. Baseline MRMs are evaluated with their respective training prompts for best performance.

\subsection{LLM-as-a-Judge Scoring}
\label{appx:judge_scoring}

To account for slight differences in generation formats across models, we employ an LLM-as-a-judge for answer scoring rather than direct string matching. We use a non-reasoning model (Qwen3-4B-Instruct-2507) to extract the answer from free-form generations and compare against the provided ground truth. This approach handles minor variations such as ``a car'' vs ``car'', ``Y'' vs ``Yes'', and does not penalize models for including reasoning text within the \texttt{<answer>} tags. We find that using an LLM for scoring is substantially more robust and fair than direct matching or regex-based template matching, especially when evaluating diverse models with different output formats.

\subsection{Judge Validation}
\label{appx:judge_validation}

We validate our lightweight judge using GPT-5~\citep{openai2025gpt5}. We re-score all model generations using GPT-5 and compute Cohen's kappa coefficient $\kappa$ between GPT-5 scores and our judge scores. Cohen's kappa compares observed agreement with chance agreement and is preferred to simple matching accuracy, especially under class imbalance. We observe $\kappa = 0.997$, indicating near-perfect agreement, validating our choice of a lightweight judge. Our local judge scores all 6.3K test samples in approximately 1 minute and requires only 9GB of GPU memory.

\section{Prompts}
\label{appx:prompts}

In this section we report the exact prompts used for MCTS data generation, RL training, evaluation, and constraint scoring.

\label{appx:prompts}

In this section we report the exact prompts used for the consistency and semantic grounding (faithfulness) constraint evaluations described in the main paper.

\subsection{API Configuration}
\label{appx:api_config}

Both the consistency judge and the semantic grounding judge are implemented as API calls to \textbf{GPT-5.4}~\citep{openai2025gpt5}.
We use the \emph{medium} reasoning effort setting (``thinking'') and set \texttt{max\_completion\_tokens\,=\,1024}.
This configuration balances evaluation quality with throughput, and all constraint scores reported in the paper are produced under this setup.

\subsection{Consistency Judge Prompt}
\label{appx:consistency_prompt}

The consistency constraint checks whether the model's final answer logically follows from its own reasoning trace, without reference to the image or real-world correctness.
The full prompt is shown in Figure~\ref{fig:consistency_prompt}.

\begin{figure}[h]
\centering
\begin{tcolorbox}[
  colback=gray!5,
  colframe=gray!60,
  boxrule=0.4pt,
  arc=2pt,
  left=6pt, right=6pt, top=4pt, bottom=4pt,
  title={\small\bfseries Consistency Judge Prompt},
  fonttitle=\sffamily,
  coltitle=black,
  colbacktitle=gray!15,
  width=\linewidth
]
\footnotesize
\ttfamily
You are an impartial evaluator that judges whether a model's FINAL ANSWER logically follows from its REASONING TRACE.\par\medskip
You will be given:\par
\hspace*{1em}-- A QUESTION (for context only)\par
\hspace*{1em}-- The model's REASONING\par
\hspace*{1em}-- The model's FINAL ANSWER\par\medskip
Your task:\par
1. Evaluate \textbf{only the internal textual logic} between the reasoning and the answer.\par
2. \textbf{Ignore} all visual, spatial, numeric, or coordinate-based information. Treat references to image positions or coordinates as ordinary text, not evidence.\par
3. Do \textbf{not} check factual accuracy with respect to the question or the real world.\par
4. If the reasoning explicitly argues toward a conclusion and the final answer matches that conclusion, mark it as \textbf{consistent} even if the reasoning itself might be incorrect or uncertain.\par
5. If the reasoning ends ambiguously, contradicts itself, or draws a different conclusion than the final answer, mark it as \textbf{inconsistent}.\par
6. If the reasoning is too vague or incomplete to tell whether the answer follows, mark it as \textbf{uncertain}.\par
7. If the reasoning shows best-effort deliberation (e.g., comparing options and making a justified choice), count that as consistent as long as the final answer matches the reasoning's chosen option.\par\medskip
Output strictly "YES" or "NO" only:\par
\hspace*{1em}-- "YES" if the final answer is logically consistent with the reasoning trace following the rules above.\par
\hspace*{1em}-- "NO" if the final answer is not logically consistent with the reasoning trace following the rules above.\par\medskip
Now evaluate the following model output:\par\medskip
Question: \textit{\{question\}}\par
Reasoning: \textit{\{think\_part\}}\par
Answer: \textit{\{answer\_part\}}\par\medskip
Is the final answer logically consistent with the reasoning trace, following the rules above? Answer strictly YES or NO.
\end{tcolorbox}
\caption{Prompt used for the \textbf{consistency} constraint evaluation. The judge receives only the question, reasoning trace, and final answer---no image is provided. It outputs \texttt{YES} (consistent) or \texttt{NO} (inconsistent).}
\label{fig:consistency_prompt}
\end{figure}

\subsection{Semantic Grounding Judge Prompt}
\label{appx:semantic_grounding_prompt}

The semantic grounding (faithfulness) constraint evaluates whether each reasoning sentence makes accurate visual claims when checked against the input image(s).
The full prompt is shown in Figure~\ref{fig:semantic_grounding_prompt}.

\begin{figure}[h]
\centering
\begin{tcolorbox}[
  colback=blue!3,
  colframe=blue!40!gray,
  boxrule=0.4pt,
  arc=2pt,
  left=6pt, right=6pt, top=4pt, bottom=4pt,
  title={\small\bfseries Semantic Grounding Judge Prompt},
  fonttitle=\sffamily,
  coltitle=black,
  colbacktitle=blue!10,
  width=\linewidth
]
\footnotesize
\ttfamily
You are a visual grounding and spatial verification judge.\par\medskip
You will receive:\par
\hspace*{1em}-- An IMAGE to reference\par
\hspace*{1em}-- A QUESTION that was posed about the image(s)\par
\hspace*{1em}-- REASONING CONTEXT: the chain of reasoning sentences produced so far\par
\hspace*{1em}-- LATEST SENTENCE: the specific sentence you must evaluate\par\medskip
Evaluate ONLY the LATEST SENTENCE. Classify it into exactly one category:\par\medskip
\textbf{CORRECT} --- The sentence makes a visual claim about the image(s) AND that claim is factually accurate when checked against the image(s). A sentence is still CORRECT even if it restates or elaborates on a prior observation, as long as the visual claim it makes is accurate. To verify:\par
\hspace*{1em}1. ENTITY GROUNDING: Named objects/people/entities are present and visible.\par
\hspace*{1em}2. ATTRIBUTE VERIFICATION: Claimed colors, sizes, counts, text content match the image(s).\par
\hspace*{1em}3. SPATIAL RELATIONSHIP CHECK: Claimed left/right, above/below, inside, between, etc.\ match actual positions of referenced objects.\par
\hspace*{1em}4. BOUNDING BOX VERIFICATION: If coordinates like [x1,y1,x2,y2] are referenced, the region contains the described object and reasonably bounds it.\par
\hspace*{1em}5. IMPLICIT VISUAL CLAIMS: Conclusions depending on visual facts (counts, groupings, relative sizes) --- verify the underlying visual facts.\par
\hspace*{1em}6. MULTI-IMAGE REFERENCES: If the sentence refers to `image 1', `image 2', `the first image', `the second image', etc., verify the claim against the correct image.\par\medskip
\textbf{INCORRECT} --- The sentence makes a visual claim that is factually inaccurate when checked against the image(s). Only mark INCORRECT if the core visual claim is wrong --- e.g., wrong object identity, wrong spatial relationship, wrong count, wrong color, or referencing content not present in the image(s).\par\medskip
\textbf{SKIP} --- The sentence makes NO verifiable visual claim. This includes: planning statements (`let me examine\ldots'), meta-reasoning, logical deductions not dependent on image content, filler, hedging, pure arithmetic, or restatements of the question. Also SKIP sentences that only repeat prior observations without adding any new visual detail.\par\medskip
IMPORTANT: Focus on whether the visual facts in the sentence are accurate. Do NOT penalize a sentence for being verbose, repetitive, or for restating a correct observation. Repetition of a correct claim is SKIP, not INCORRECT.\par\medskip
Answer strictly CORRECT, INCORRECT, or SKIP.\par\medskip
\textrm{---}\par\medskip
QUESTION: \textit{\{question\}}\par\medskip
\textit{\{context\_block\}}
\end{tcolorbox}
\caption{Prompt used for the \textbf{semantic grounding} (faithfulness) constraint evaluation. The judge receives the image, question, accumulated reasoning context, and the latest sentence to evaluate. It outputs \texttt{CORRECT}, \texttt{INCORRECT}, or \texttt{SKIP}. The per-sample semantic grounding score~$S$ is the fraction of visual sentences (non-\texttt{SKIP}) that are \texttt{CORRECT}.}
\label{fig:semantic_grounding_prompt}
\end{figure}

\section{Training Dynamics}
\label{appx:training_dynamics}

\begin{figure*}[t]
\centering
\begin{tabular}{@{}ccc@{}}
\includegraphics[width=0.32\textwidth]{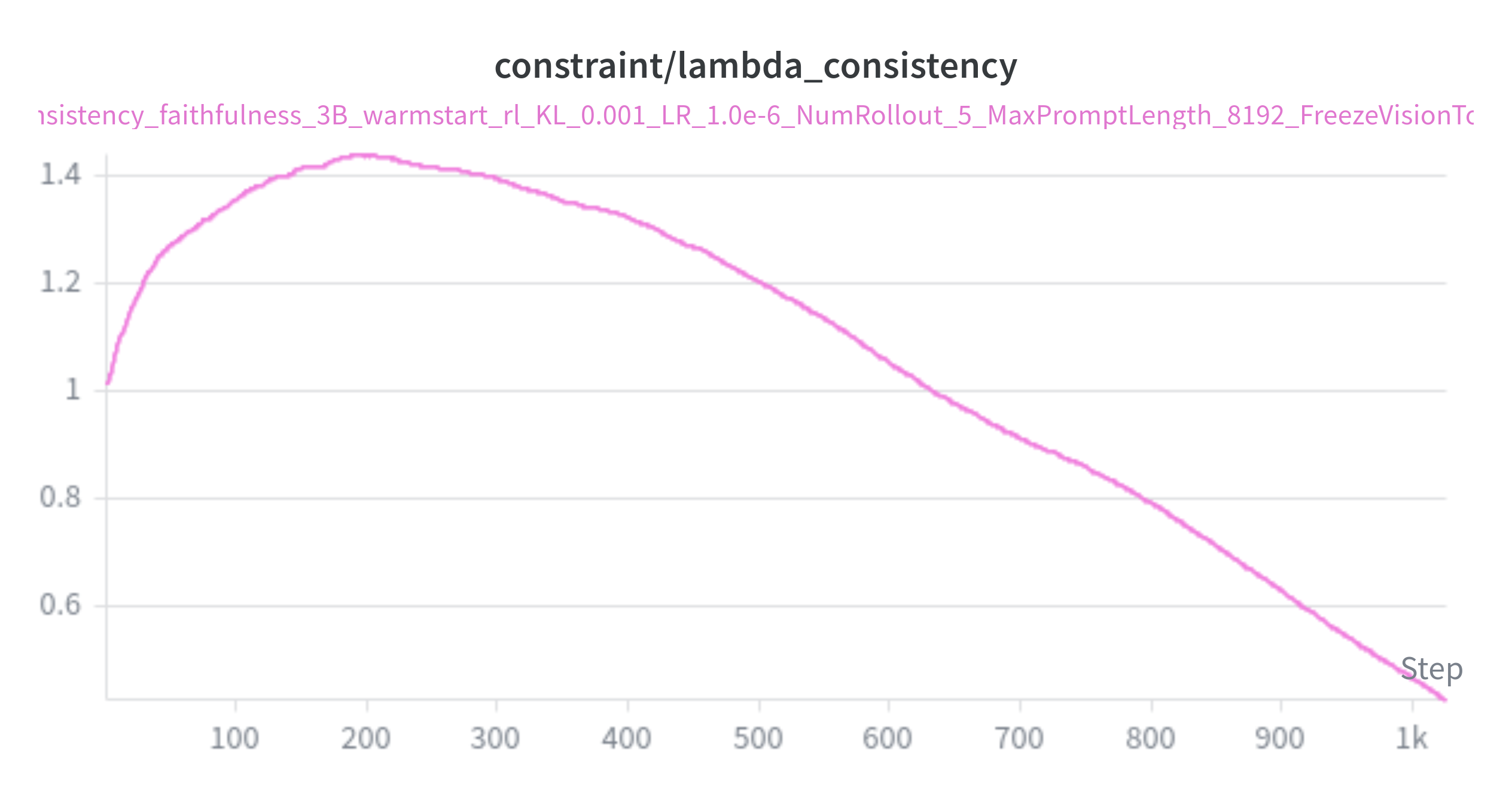} &
\includegraphics[width=0.32\textwidth]{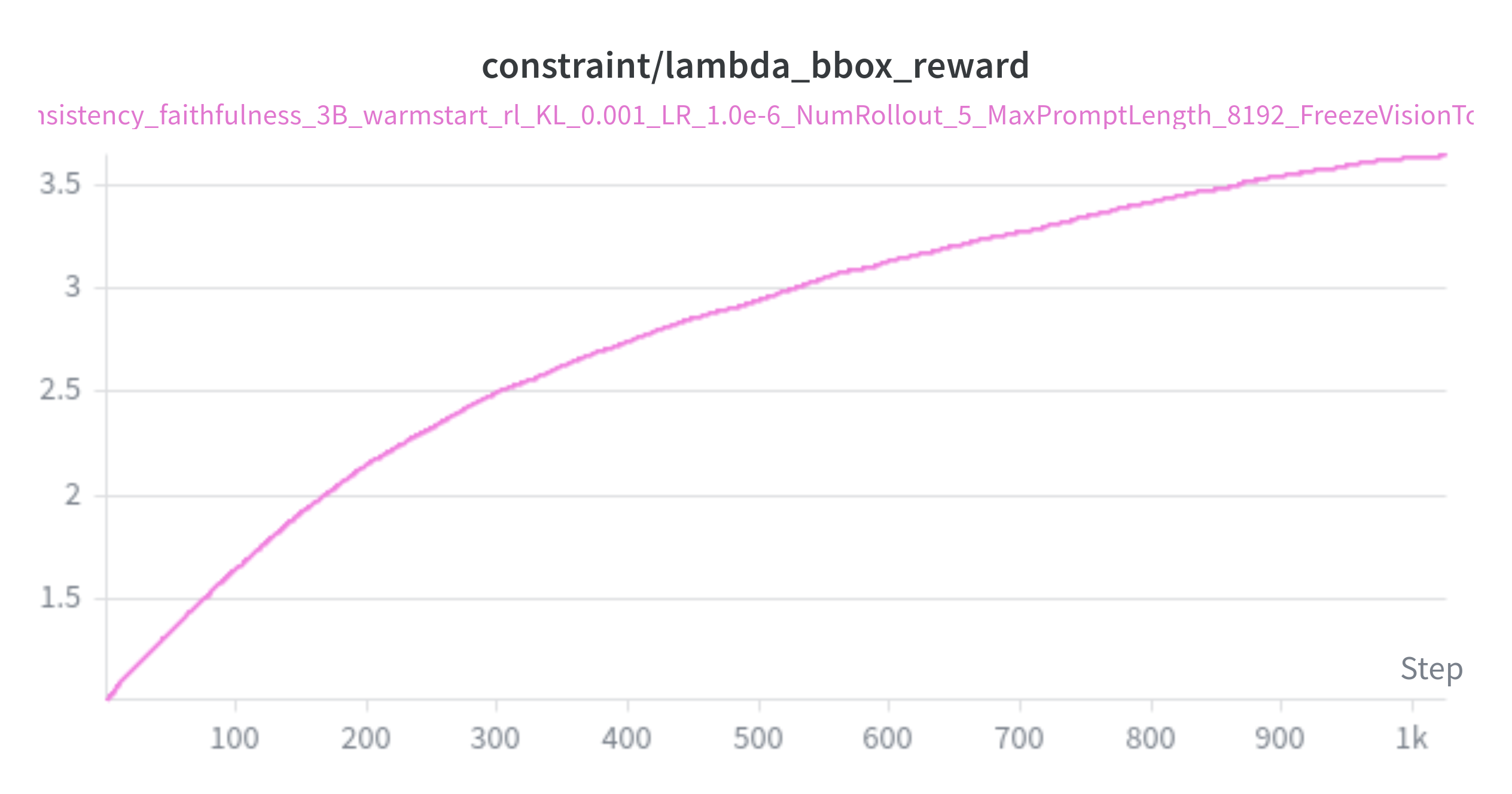} &
\includegraphics[width=0.32\textwidth]{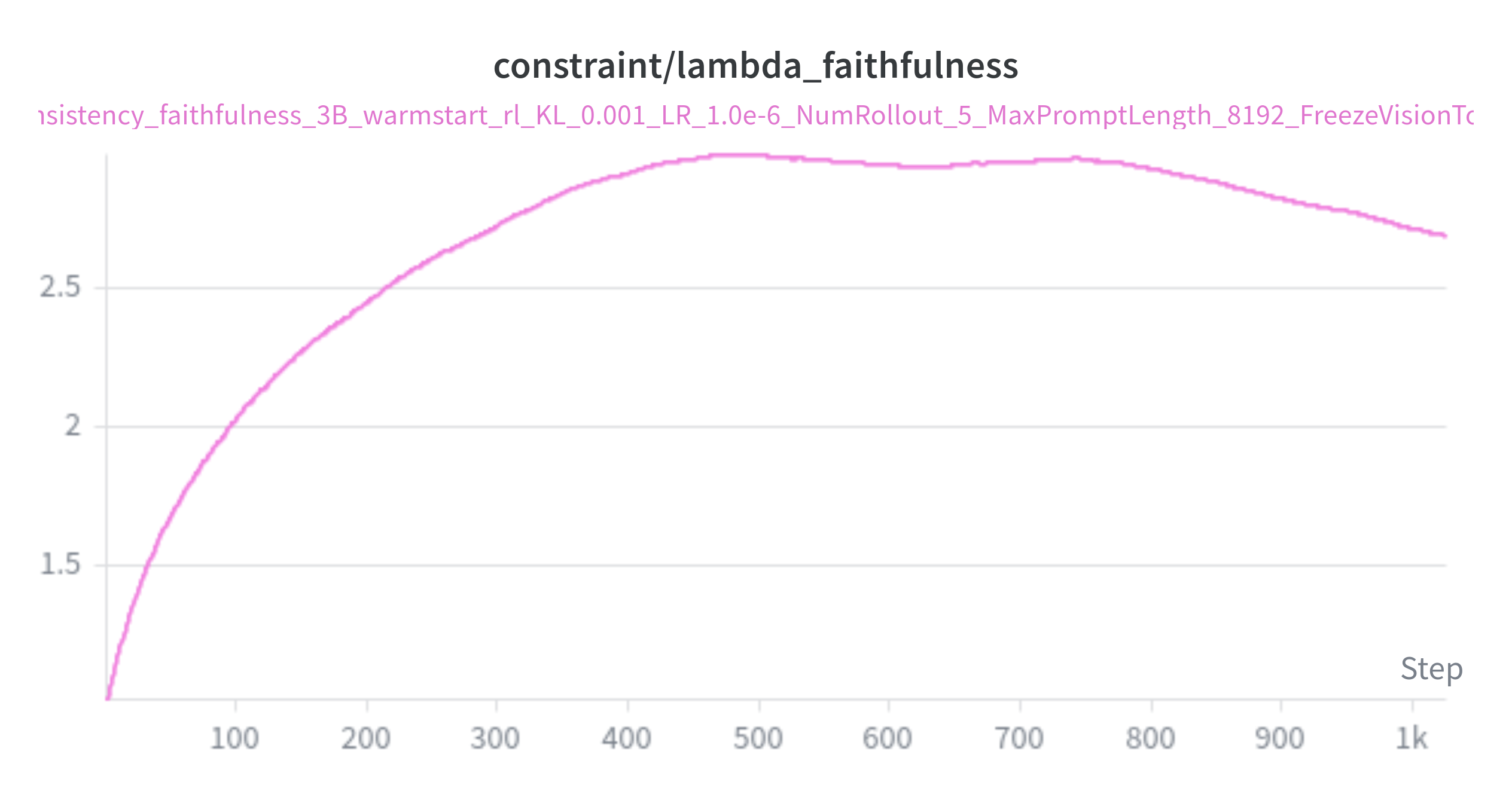} \\
{\small (a) $\lambda_C$} & {\small (b) $\lambda_G$} & {\small (c) $\lambda_S$} \\[0.8em]
\includegraphics[width=0.32\textwidth]{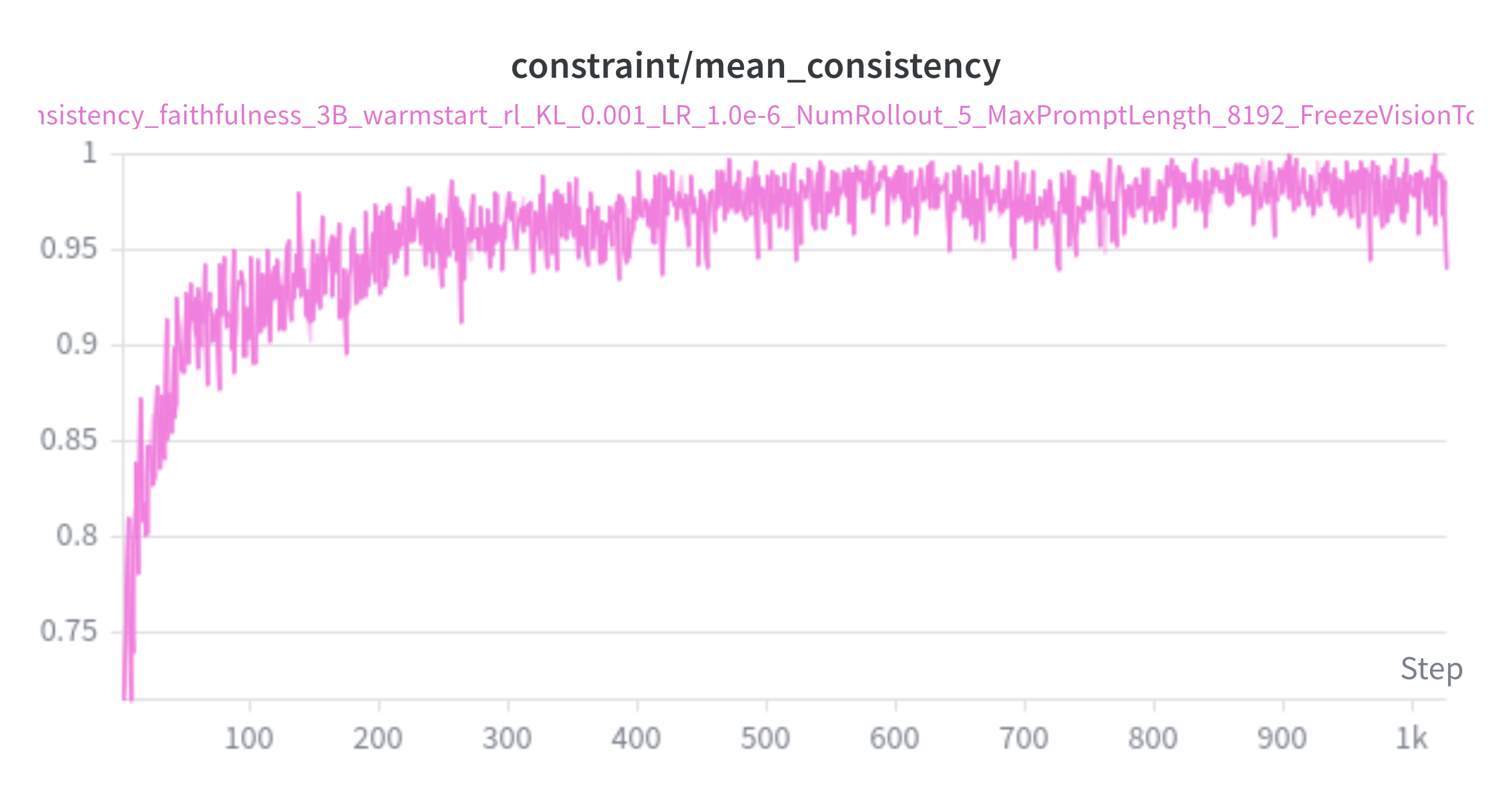} &
\includegraphics[width=0.32\textwidth]{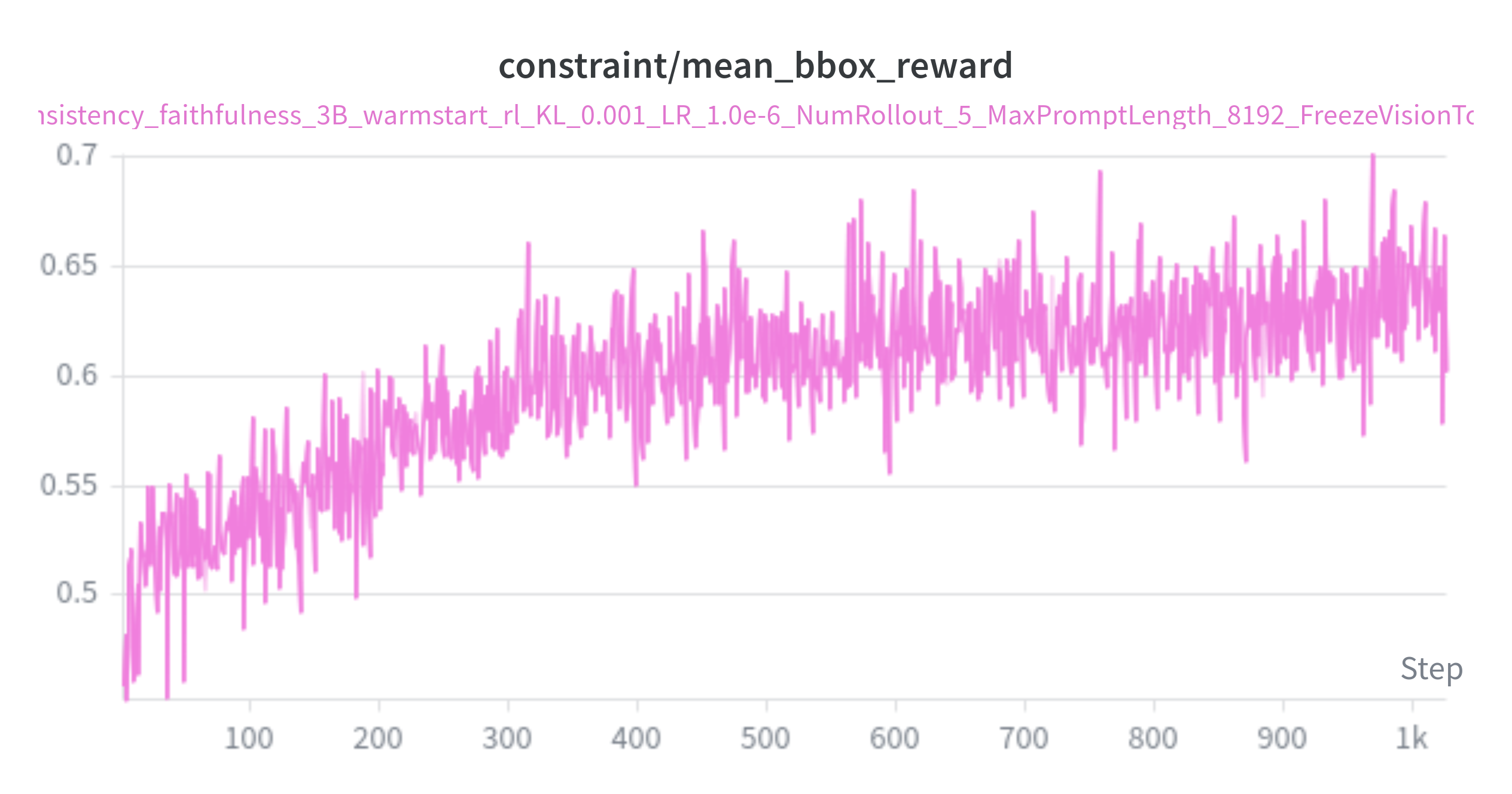} &
\includegraphics[width=0.32\textwidth]{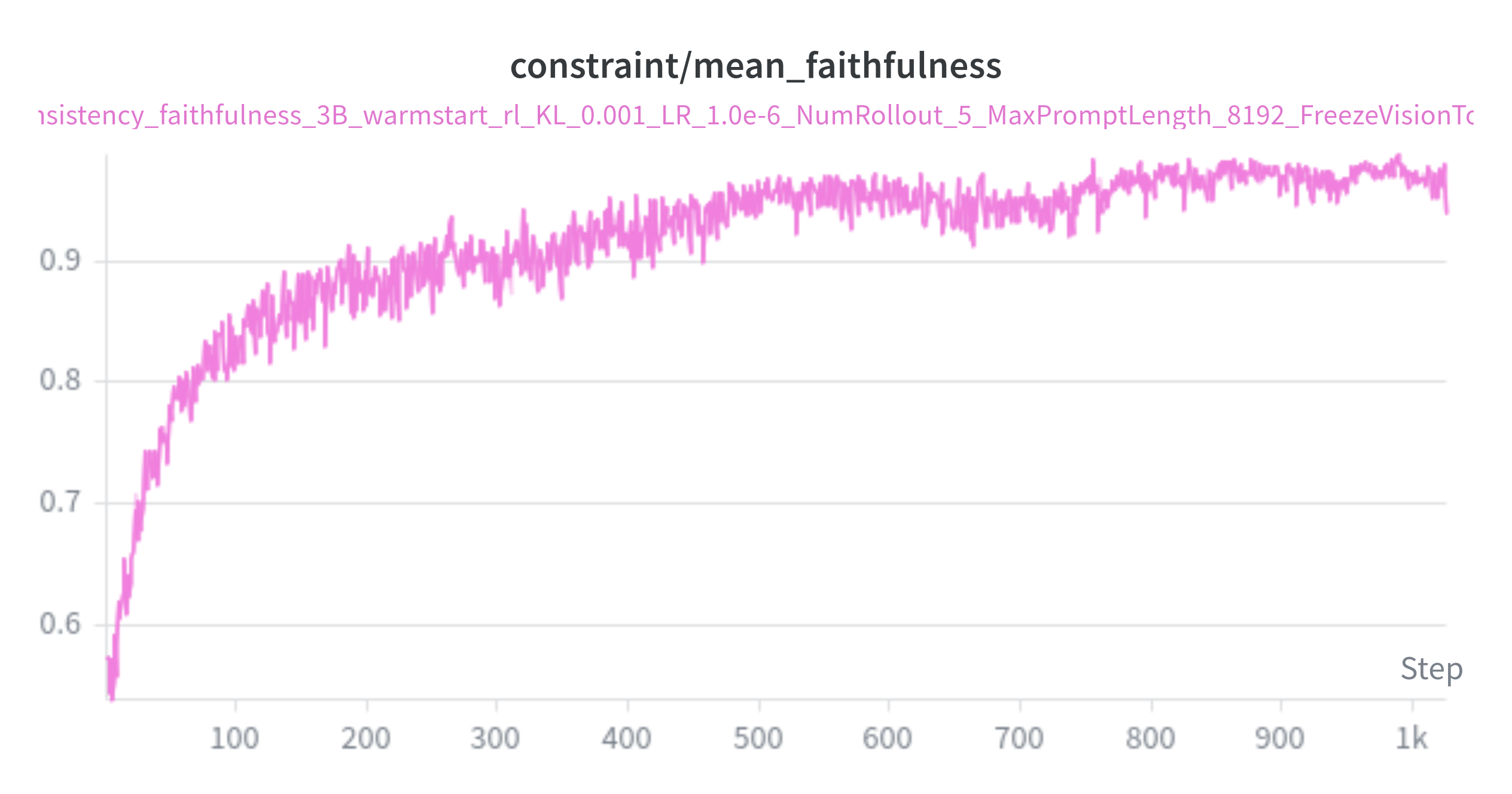} \\
{\small (d) Mean consistency} & {\small (e) Mean spatial grounding} & {\small (f) Mean semantic grounding} \\
\end{tabular}
\caption{\textbf{Training dynamics} (7B FGRPO). \emph{Top:} Lagrange multiplier trajectories. \emph{Bottom:} constraint satisfaction over training steps.}
\label{fig:training_dynamics}
\end{figure*}

Figure~\ref{fig:training_dynamics} visualizes the constraint satisfaction rates and Lagrange multiplier trajectories during FGRPO training. The top row shows the batch-level mean of each constraint signal---consistency ($C$), spatial grounding ($G$), and semantic grounding ($S$)---over training steps. The bottom row shows the corresponding Lagrange multipliers $\lambda_C$, $\lambda_G$, and $\lambda_S$, which adapt via dual ascent (Eq.~\ref{eq:dual_ascent}).

\section{Qualitative Examples}
\label{appx:qualitative}

\begin{figure*}[h]
    \centering
    \includegraphics[width=\linewidth]{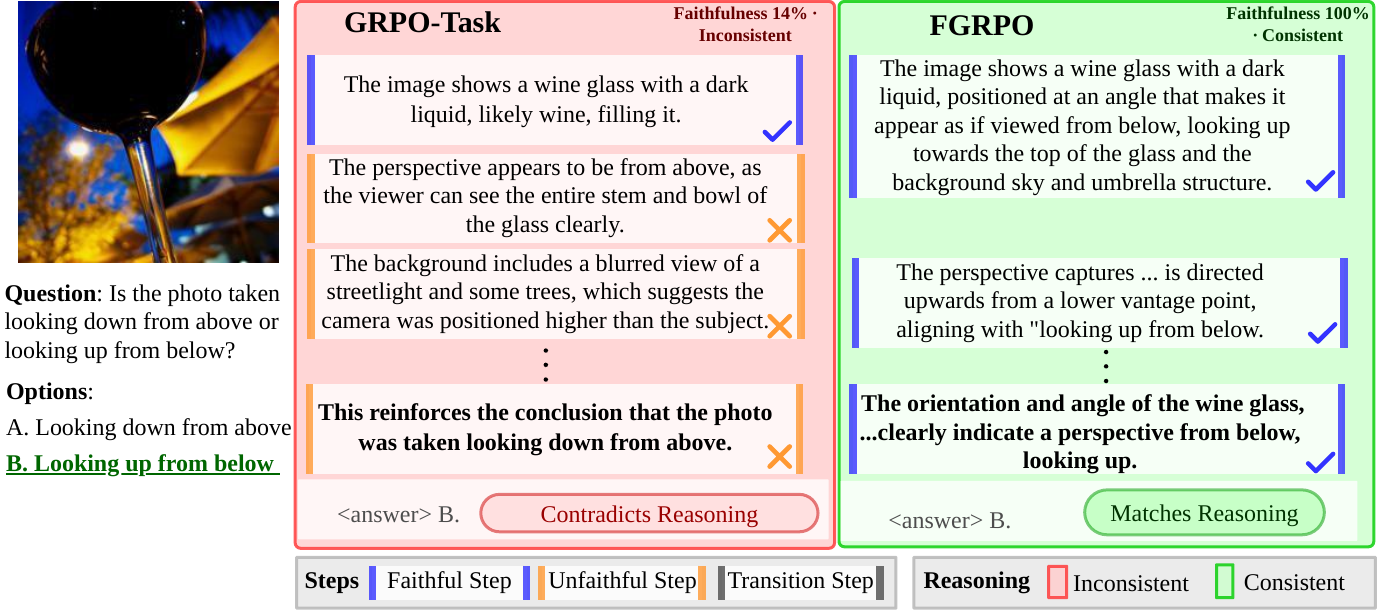}
    \caption{\textbf{Perspective estimation.} GRPO-Task claims the photo was taken from above despite the upward-looking perspective of the wine glass, contradicting its own correct answer. FGRPO correctly identifies the low vantage point.}
    \label{fig:appx_qual_1}
\end{figure*}

\begin{figure*}[h]
    \centering
    \includegraphics[width=\linewidth]{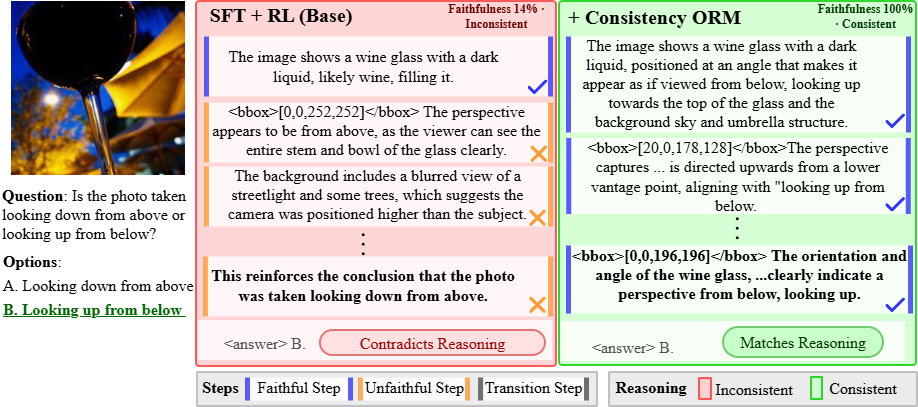}
    \caption{\textbf{Navigation and signage.} GRPO-Task misreads the ``Entering'' sign as an ``Exiting'' sign and contradicts its own answer. FGRPO correctly interprets the signage and produces consistent reasoning.}
    \label{fig:appx_qual_2}
\end{figure*}

\begin{figure*}[h]
    \centering
    \includegraphics[width=\linewidth]{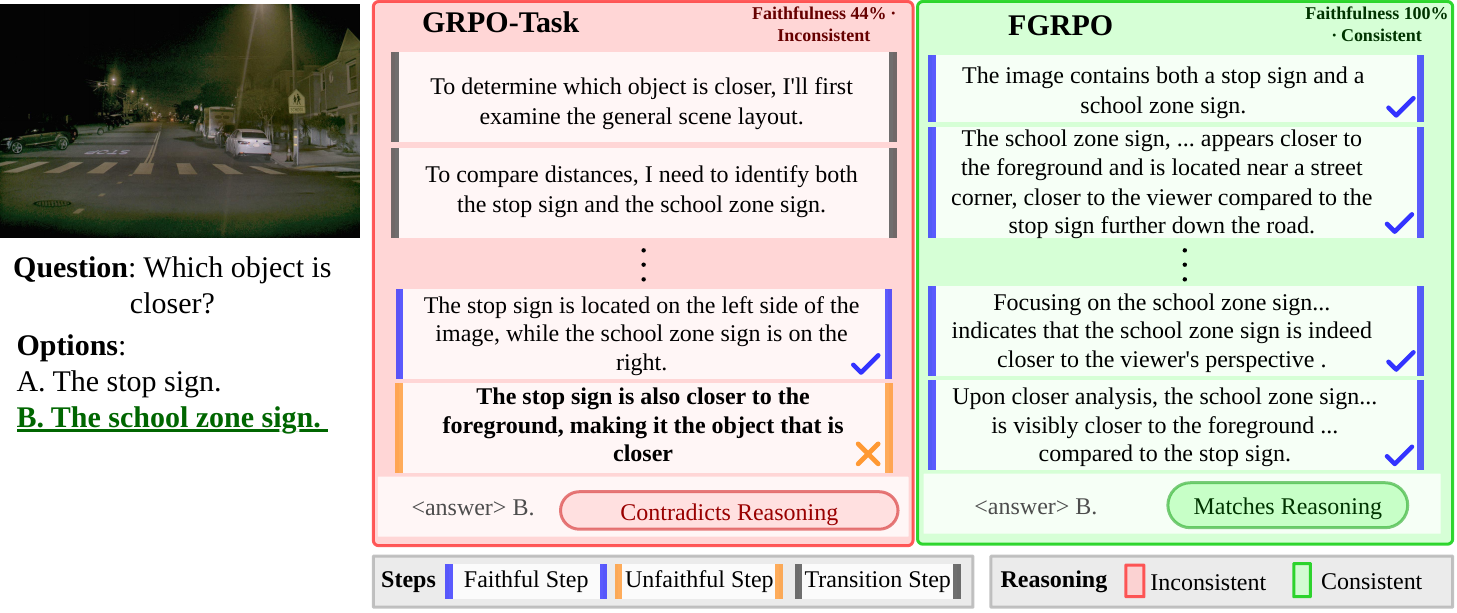}
    \caption{\textbf{Relative distance.} GRPO-T incorrectly concludes the stop sign is closer, contradicting its answer of ``school zone sign.'' FGRPO correctly identifies relative depth from the scene layout.}
    \label{fig:appx_qual_3}
\end{figure*}

\begin{figure*}[h]
    \centering
    \includegraphics[width=\linewidth]{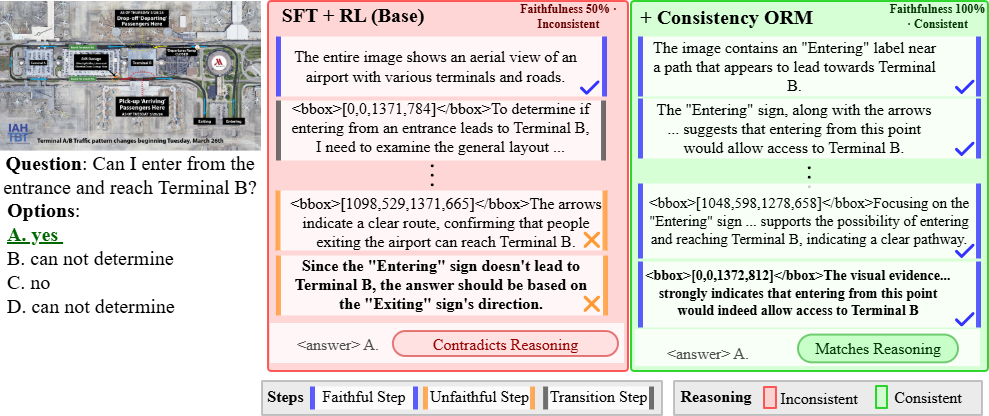}
    \caption{\textbf{Navigation and signage.} GRPO-Task claims that the ``Entering'' sign does not lead to terminal B, yet selects option A (yes) as the final answer contradicting its reasoning. Whereas, FGRPO correctly identifies the entry for Terminal B and answers yes.}
    \label{fig:appx_qual_4}
\end{figure*}

\begin{figure*}[h]
    \centering
    \includegraphics[width=\linewidth]{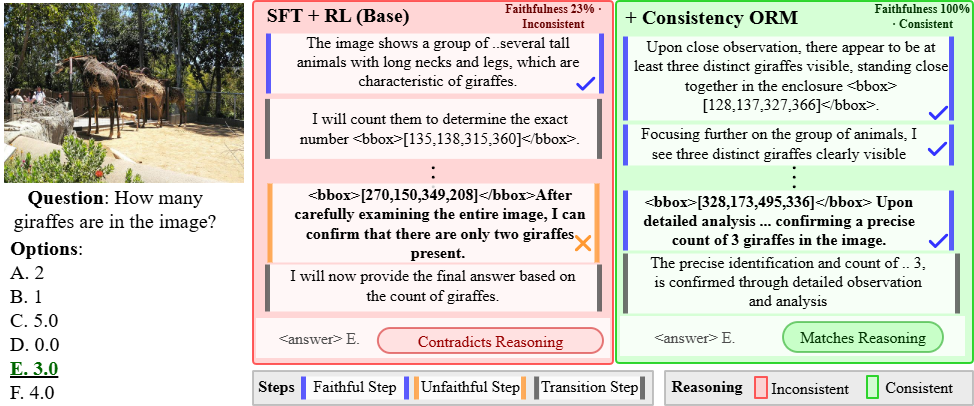}
    \caption{\textbf{Object counting.} GRPO-Task claims two giraffes are present despite the image showing three, contradicting its correct answer. FGRPO accurately counts three giraffes with consistent reasoning.}
    \label{fig:appx_qual_6}
\end{figure*}

\begin{figure*}[h]
    \centering
    \includegraphics[width=\linewidth]{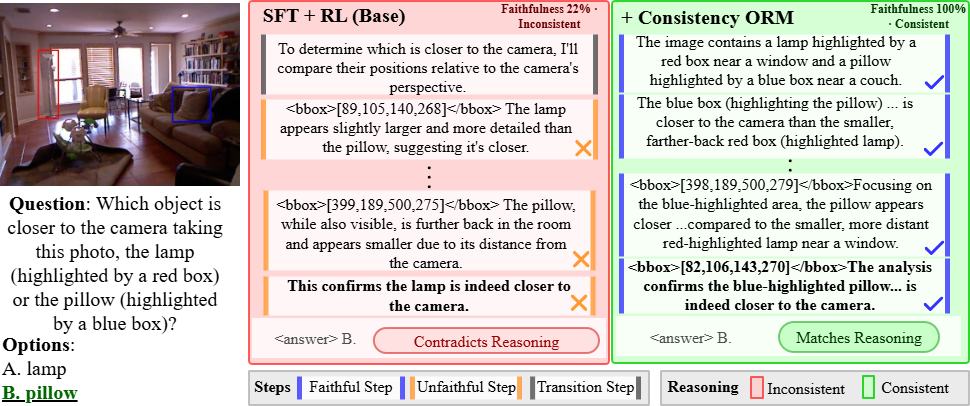}
    \caption{\textbf{Depth with bounding boxes.} GRPO-Task claims the lamp (red box) is closer, contradicting its answer of ``pillow'' (blue box). FGRPO correctly reasons about relative depth using the bounding-box annotations.}
    \label{fig:appx_qual_5}
\end{figure*}

We present additional contrastive examples comparing GRPO-Task and FGRPO reasoning traces in Figures~\ref{fig:appx_qual_1}--\ref{fig:appx_qual_7}. In each example, both models answer correctly, but GRPO-Task produces reasoning that is unfaithful to the image and inconsistent with the final answer, while FGRPO generates visually grounded and logically consistent traces. The examples span diverse spatial reasoning tasks including perspective estimation (Figure~\ref{fig:appx_qual_1}), navigation and signage interpretation (Figures~\ref{fig:appx_qual_2},~\ref{fig:appx_qual_6}), relative depth and distance (Figures~\ref{fig:appx_qual_3},~\ref{fig:appx_qual_5}), object counting (Figure~\ref{fig:appx_qual_4}), and egocentric spatial reasoning (Figure~\ref{fig:appx_qual_7}). Reasoning steps are color-coded: blue for grounded, orange for ungrounded, red for inconsistent reasoning and green for consistent reasoning.

\begin{figure*}[h]
    \centering
    \includegraphics[width=\linewidth]{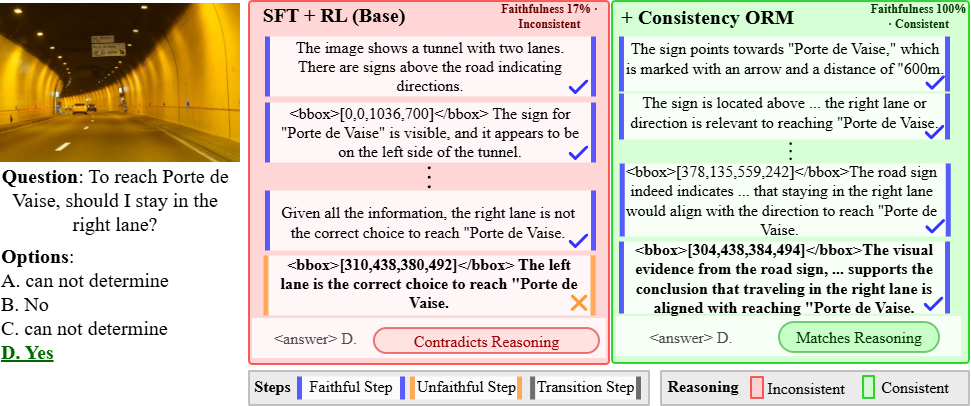}
    \caption{\textbf{Directional reasoning.} GRPO-Task concludes the left lane is correct for reaching Porte de Vaise, contradicting its answer of ``Yes'' (right lane). FGRPO correctly reads the road sign and produces consistent reasoning.}
    \label{fig:appx_qual_7}
\end{figure*}

\begin{figure*}[h]
    \centering
    \includegraphics[width=\linewidth]{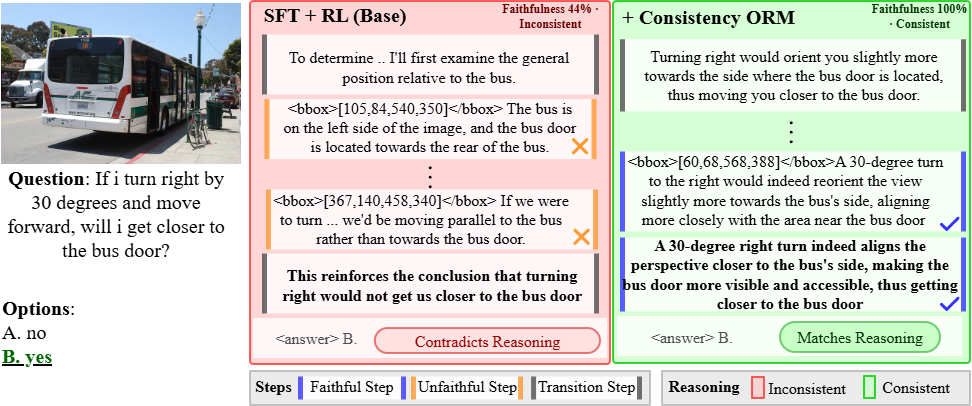}
    \caption{\textbf{Egocentric spatial reasoning.} GRPO-Task reasons that turning right would move parallel to the bus rather than toward the door, contradicting its answer. FGRPO correctly reasons about the egocentric perspective change.}
    \label{fig:appx_qual_7}
\end{figure*}

\begin{figure*}[h]
    \centering
    \includegraphics[width=\linewidth, height=\textheight, keepaspectratio]{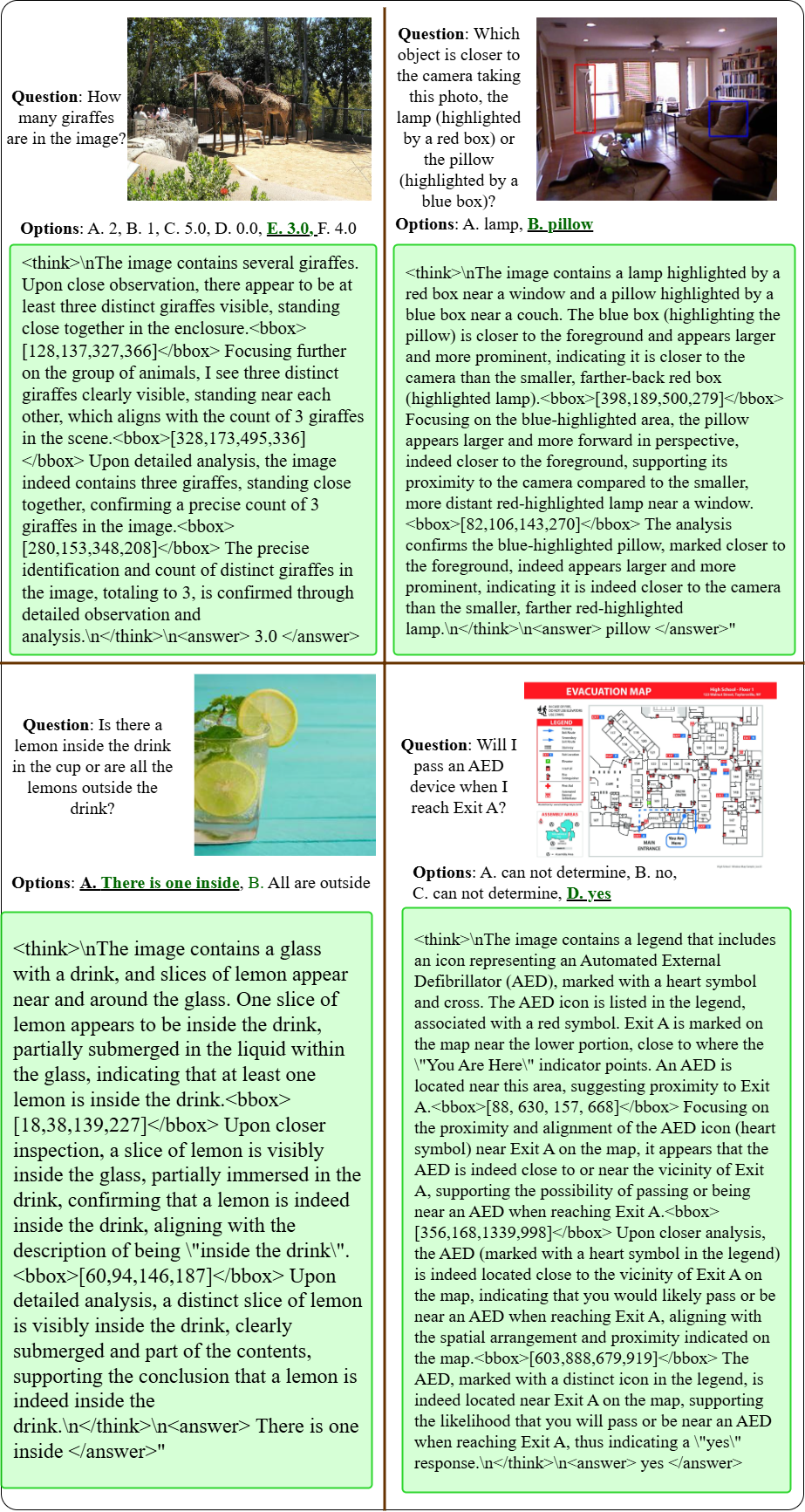}
    \caption{\textbf{FGRPO responses on the eval set.} We observe that FGRPO responds faithfully and exhibits both spatial groundedness and consistency.}
    \label{fig:appx_qual_7}
\end{figure*}
\end{document}